%% file: arxiv.tex
\RequirePackage{fix-cm}
\documentclass{article}
\usepackage{arxiv}
\renewcommand{\headeright}{}
\renewcommand{\shorttitle}{}
\usepackage[round,authoryear]{natbib}
\usepackage{hyperref}
\usepackage{url}
\usepackage[all]{hypcap}

\usepackage{amsmath}
\usepackage[capitalize,noabbrev]{cleveref}
\hypersetup{
    colorlinks = true,
    citecolor = {magenta},
}

\usepackage[nopatch=footnote]{microtype}
\usepackage{graphicx}
\usepackage{subfigure}
\usepackage{booktabs} 

\usepackage{xcolor}
\usepackage{colortbl}
\usepackage{algorithm}
\usepackage{algpseudocode}
\usepackage{setspace}

\usepackage{amssymb}
\usepackage{mathtools}
\usepackage{amsthm}

\usepackage{placeins}

\usepackage{tabularx}

\usepackage{float}

\usepackage[most,skins,theorems]{tcolorbox}
\tcbset{
  aibox/.style={
    width=\linewidth,
    top=7pt,
    bottom=2pt,
    colback=rliableolive!13!white,
    colframe=black,
    colbacktitle=black,
    enhanced,
    center,
    attach boxed title to top left={yshift=-0.1in,xshift=0.15in},
    boxed title style={boxrule=0pt,colframe=white,},
  }
}
\newtcolorbox{AIbox}[2][]{aibox,title=#2,#1}

\usepackage{wrapfig}
\definecolor{lightblue}{rgb}{0.22,0.45,0.70}

\definecolor{rliableolive}{HTML}{BBCC33}
\definecolor{rliableblue}{HTML}{77AADD}
\definecolor{rliablered}{HTML}{EE8866}

\usepackage{crossreftools}
\usepackage[suppress]{color-edits}

\usepackage{dsfont}
\usepackage{nicefrac}
\newtcolorbox{analysisbox}[1][]{
    enhanced jigsaw,
    colback=white,
    colframe=blue!75!black,
    fonttitle=\bfseries,
    boxsep=5pt,
    left=5pt,
    right=5pt,
    top=5pt,
    bottom=5pt,
    title=#1,
}
\definecolor{editInitialResponse}{RGB}{255, 235, 156} 
\definecolor{editBacktrack}{RGB}{0, 0, 139}
\definecolor{editRevisedResponse}{RGB}{255, 182, 193}

\usepackage{pifont}
\usepackage{soul}
\definecolor{highlightmistake}{RGB}{255, 179, 179} 
\definecolor{highlightcorrect}{RGB}{179, 255, 179}

\usepackage[capitalize,noabbrev]{cleveref}

\theoremstyle{plain}

\theoremstyle{definition}

\theoremstyle{remark}

\usepackage{multirow}
\usepackage{enumitem}
\input{math_commands}

\definecolor{darkblue}{rgb}{0, 0, 0.5}
\definecolor{promptbg}{RGB}{232,242,255}
\definecolor{promptframe}{RGB}{92,140,214}
\definecolor{evalbg}{RGB}{241,252,241}
\definecolor{evalframe}{RGB}{77,162,94}
\definecolor{solutionbg}{RGB}{255,246,232}
\definecolor{solutionframe}{RGB}{214,139,42}
\hypersetup{colorlinks=true, citecolor=darkblue, linkcolor=darkblue, urlcolor=darkblue}

\newtcolorbox{promptbox}[1]{
  enhanced,
  breakable,
  arc=2.2mm,
  boxrule=0.8pt,
  left=1.3mm,right=1.3mm,top=1.0mm,bottom=1.1mm,
  colback=promptbg,
  colframe=promptframe,
  colbacktitle=promptframe!34,
  coltitle=black,
  fonttitle=\bfseries,
  title={#1}
}

\newtcolorbox{evalbox}[1]{
  enhanced,
  breakable,
  arc=2.2mm,
  boxrule=0.8pt,
  left=1.3mm,right=1.3mm,top=1.0mm,bottom=1.1mm,
  colback=evalbg,
  colframe=evalframe,
  colbacktitle=evalframe!30,
  coltitle=black,
  fonttitle=\bfseries,
  title={#1}
}

\newtcolorbox{smallevalbox}[1]{
  enhanced,
  breakable,
  arc=2.2mm,
  boxrule=0.8pt,
  left=1.3mm,right=1.3mm,top=1.0mm,bottom=1.1mm,
  colback=evalbg,
  colframe=evalframe,
  colbacktitle=evalframe!30,
  coltitle=black,
  fonttitle=\bfseries,
  title={#1}
}

\newtcolorbox{solutionbox}[1]{
  enhanced,
  breakable,
  arc=2.2mm,
  boxrule=0.8pt,
  left=1.3mm,right=1.3mm,top=1.0mm,bottom=1.1mm,
  colback=solutionbg,
  colframe=solutionframe,
  colbacktitle=solutionframe!34,
  coltitle=black,
  fonttitle=\bfseries,
  title={#1}
}
\usepackage{xcolor}
\usepackage{listings}
\usepackage{textcomp}
\usepackage{upquote}
\lstdefinestyle{mypython}{
  language=Python,
  basicstyle=\ttfamily\small,
  keywordstyle=\color{blue!70!black}\bfseries,
  stringstyle=\color{green!45!black},
  commentstyle=\color{gray}\itshape,
  showstringspaces=false,
  breaklines=true,
  breakatwhitespace=true,
  columns=fullflexible,
  frame=single,
  framesep=4pt,
  xleftmargin=10pt,
  literate={`}{{\textasciigrave}}1
}

\newcommand{\cmark}{\textcolor{green!50!black}{\ding{51}}}
\newcommand{\xmark}{\textcolor{red!75!black}{\ding{55}}}

\newcommand{\methodname}{\texttt{MIRROR}}
\newcommand{\methodfullname}{\textbf{Modality-Informed Reciprocal Reasoning Optimization}}
\newcommand{\mirrorRKLEquationWidth}{0.78\linewidth}

\title{MIRROR: Learning from the Other View for Multi-Modal Reasoning}

\author{%
  Wen Ye$^1$\thanks{Equal contribution. Corresponding Author: yewen@usc.edu, yuxiaoq@andrew.cmu.edu}, Yuxiao Qu$^2$\protect\footnotemark[1], Aviral Kumar$^2$, Xuezhe Ma$^1$ \\
  $^1$University of Southern California, $^2$Carnegie Mellon University \\
  {\small\normalfont\texttt{yewen@usc.edu, yuxiaoq@andrew.cmu.edu}} \\
  {\small\normalfont\texttt{aviralku@andrew.cmu.edu, xuezhema@usc.edu}}
}
\date{}

\begin{document}

\maketitle

\input{sections/abstract}

\input{sections/introduction}

\input{sections/relwork}

\input{sections/pilot}

\input{sections/methodology}

\input{sections/experiment}

\input{sections/conclusion_arxiv}
\input{sections/acknowledgements}

\bibliography{main}

\newpage
\appendix
\part*{Appendices}
\input{sections/app_additional_results}
\input{sections/app_error_analysis}
\input{sections/app_computation}
\input{sections/app_pseudocode}
\input{sections/app_hyperparameters}

\input{sections/app_prompt}
\input{sections/app_examples}

\clearpage
\newpage

\end{document}

%% file: math_commands.tex
\usepackage{amsmath,amsfonts,bm}

\def\eqref#1{Eq.~\ref{#1}}

\def\1{\bm{1}}

\DeclareMathAlphabet{\mathsfit}{\encodingdefault}{\sfdefault}{m}{sl}
\SetMathAlphabet{\mathsfit}{bold}{\encodingdefault}{\sfdefault}{bx}{n}

\newcommand{\by}{\mathbf{y}}

\newcommand{\bx}{\mathbf{x}}


%% file: sections/abstract.tex
\begin{abstract}
Vision-language models (VLMs) can behave very differently depending on whether the same underlying problem is presented textually or visually. This inconsistency suggests that different views expose different reasoning paths and failure modes that standard multimodal post-training does not fully exploit. To study and exploit this phenomenon, we construct ODA-Data, a high-quality paired multimodal geometry dataset with text-dominant, image-dominant, and combined image+text views of the same problems, together with splits for training and evaluating view-dependent reasoning behaviors. We then develop \methodfullname{} (\methodname{}), a reinforcement learning approach that derives a cross-view self-distillation signal from the model’s own modality-dependent capabilities. For each problem, \methodname{} evaluates the model under all views, selects the best-performing view as a teacher, and trains the remaining views using a GRPO task-reward objective augmented with a reverse-KL regularizer toward that teacher. Across reasoning benchmarks that evaluate on multimodal mathematics problems, \methodname{} improves over standard RL and yields more accurate and consistent behavior across different input modality representations.
\end{abstract}

%% file: sections/introduction.tex
\section{Introduction}
\label{sec:intro}

Vision-language models have made substantial progress on multimodal reasoning~\citep{chen2024internvlscalingvisionfoundation,bai2023qwenvlversatilevisionlanguagemodel}, but their behavior remains highly sensitive to how the information is presented~\citep{li2025surveystateartlarge,ismithdeen2025promptceptionsensitivelargemultimodal}.
As we also find in our experiments, a model may solve a problem when the relevant information is written in text, yet fail when those relations must be recovered from an image; conversely, an image may make geometric structure salient in a way that a purely textual description does not. This inconsistency is especially concerning because the underlying reasoning problem has not changed, only the input view has. It suggests that current VLMs are not merely limited by problem difficulty~\citep{yao2025lensmultilevelevaluationmultimodal}, but also by view-dependent failures in grounding and reasoning. Our goal is to study and address these failures.

To do so, we focus on geometry problems that serve as a natural testbed for studying this phenomenon~\citep{zhang2024mathversedoesmultimodalllm}. Many geometry problems can be presented through several views associated with the same underlying problem: a \emph{text-dominant} view, where quantities and relations are enumerated in language; an \emph{image-dominant} view, where the image carries essential geometric information and the accompanying text is minimal and mainly specifies the question; and a \emph{combined} image+text view, where both sources are available. Since they share the same problem identity and answer target, they enable a controlled comparison of view-dependent reasoning. As we find in our initial study (Table~\ref{tab:modality_examples_compact}), when the same model is asked to solve the same underlying problem, one input view can yield correct reasoning substantially more reliably than another. Such view-dependent behavior is not merely a robustness failure; it can also provide a useful supervision signal.

Prior work on multimodal reasoning often improves models by using external teachers~\citep{liu2023visualinstructiontuning}, curated rationales~\citep{zhang2024multimodalchainofthoughtreasoninglanguage}, or specialized training data that teaches the model how to combine visual and textual inputs~\citep{dong2025insightvexploringlongchainvisual,bousselham2025voldreasoningtransferllms}. However, such supervision may not be available for every target setting, and distillation from external teachers can be fragile under distribution mismatch~\citep{li2026rethinking}. In contrast, we study whether paired views of the same problem can provide a form of self-supervision: when the model performs substantially better under one view than another, can the stronger view help improve the weaker one?

\begin{figure}[t]
\centering

\includegraphics[width=0.99\linewidth]{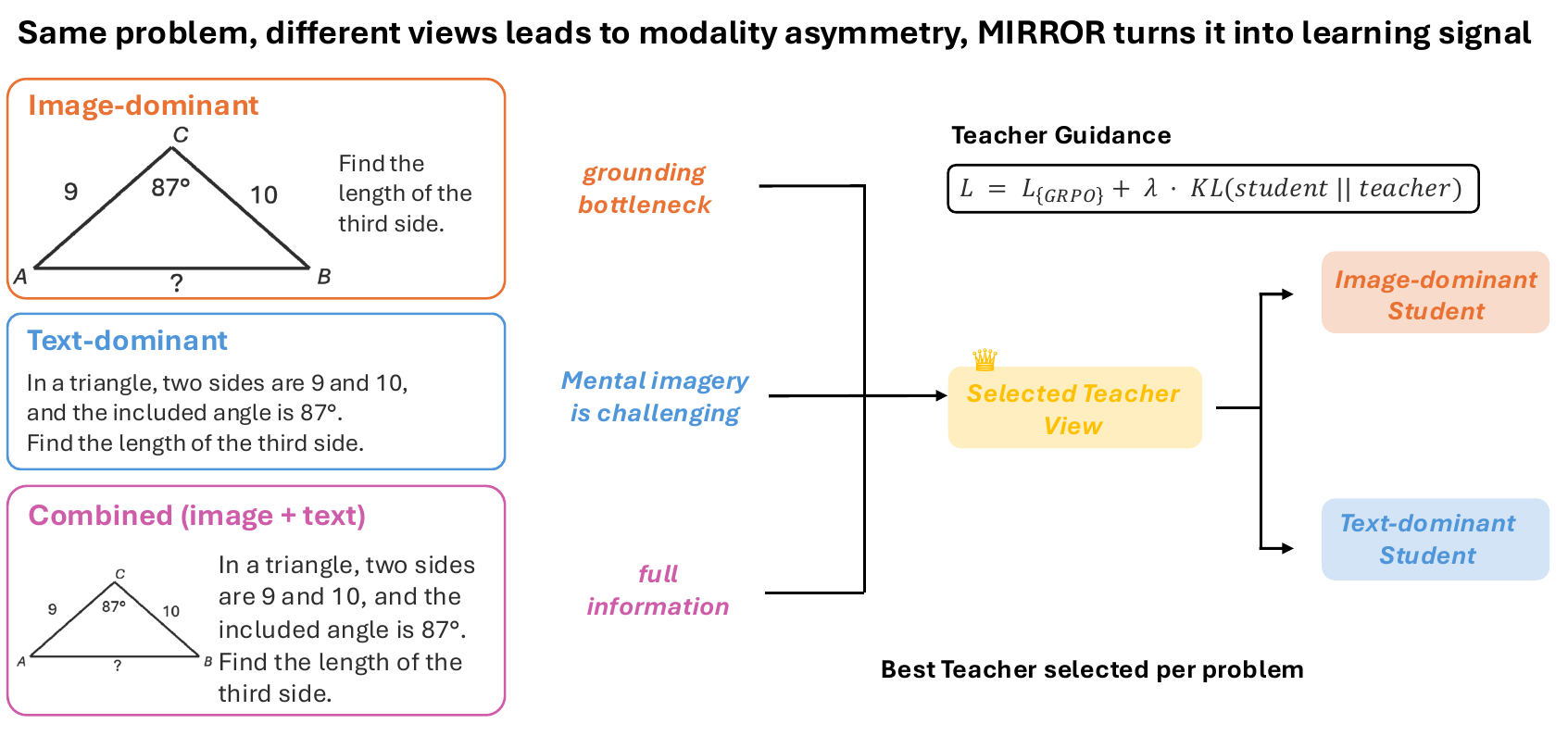}
\caption{\footnotesize{
\textbf{\methodfullname{} (\methodname{}).}
\methodname{} exploits view asymmetry by selecting the strongest-performing view of each problem as a teacher and regularizes weaker student views towards the teacher distribution, improving performance without external supervision.
}}
\label{fig:method_overview}

\end{figure}

In this paper, we propose \methodname{} (\methodfullname{}), a training framework that uses reasoning from stronger views of a problem to improve reasoning under weaker views. For each problem, we construct text-dominant, image-dominant, and combined image+text views (Figure~\ref{fig:method_overview}), evaluate the current model under each view, and select the strongest-performing view as a problem-specific teacher. The student is trained under either text-dominant or image-dominant which are restricted views with no information redundancy. The challenge is to transfer the teacher signal without forcing the student to imitate trajectories that depend on information unavailable from its own view. \methodname{} addresses this with on-policy distillation: the student samples rollouts from its own view, while the selected teacher view only rescores the same student-generated trajectories. This on-policy distillation provides dense, token-level teacher guidance on states actually visited by the student. We combine it with outcome-reward RL because the selected teacher view is only relatively stronger and may still be imperfect; the outcome reward anchors learning to final-answer correctness rather than teacher imitation alone.

To instantiate our approach, we construct ODA-Data, a paired geometry dataset with text-dominant, image-dominant, and combined image+text views of the same underlying problems.  ODA-Data preserves problem identity across views, enabling us to measure both accuracy and cross-view consistency. We use ODA-Train for post-training and ODA-Val as a held-out benchmark to measure view-dependent solvability: whether a model solves the same problem consistently when presented through different views. Empirically, \methodname{} outperforms both post-training on a mixture of prompts and recent open-source post-trained VLMs. On ODA-Val, \methodname{} achieves the best performance across all image and text validation metrics, improving image pass@16 by 14.49 percentage points (pp) and text pass@16 by 5.88 pp over the base model. These gains transfer to external benchmarks: on GeoInt, \methodname{} improves over the base model by 7.59 pp in pass@16, and on MathVerse, it improves the mean score by 5.56 points. Beyond accuracy, on ODA-Train, the fraction of problems solved under both image-dominant and text-dominant views increases from 42.5\% to 60.7\% with \methodname{} training. \methodname{} also expands the set of solvable examples, achieving the largest net solvability gain on ODA-Data at +8.95 pp. These results show that paired views provide an effective self-supervision signal for stronger and more consistent multimodal reasoning.

\vspace{-0.2cm}

%% file: sections/relwork.tex
\section{Related Work}
\label{sec:related}

\textbf{Multimodal mathematical reasoning benchmarks and data scaling.}
Prior work shows that specialized multimodal math data can substantially improve MLLM reasoning. For example, Math-LLaVA fine-tunes LLaVA-1.5 on MathV360K~\citep{shi2024mathllavabootstrappingmathematicalreasoning}, and MultiMath-7B trains on MultiMath-300K with image captions and step-by-step solutions~\citep{peng2024multimathbridgingvisualmathematical}. Several benchmarks have also been introduced for visual mathematical reasoning, including MATH-Vision~\citep{wang2024measuringmultimodalmathematicalreasoning}, MathVerse~\citep{zhang2024mathversedoesmultimodalllm}, and geometry-focused benchmarks such as LeanGeo-Bench~\citep{song2025leangeoformalizingcompetitionalgeometry}, GeoInt~\citep{wei2025geointr1formalizingmultimodalgeometric}, and GeoGoal~\citep{chen2026milestonesoutcomeunlockinggeometric}. More recently, CROSSMATH studies modality gaps using matched text-dominant, image-dominant, and the combined image+text versions of synthetic puzzles~\citep{xu2026visionlanguagemodelstrulyperform}. These works motivate our focus on geometry and provide evaluation settings that we build on, including MathVerse and GeoInt. Instead of focusing on scaling data or constructing benchmarks, our goal is algorithmic: how paired views of the same problem can be used during RL training to transfer reasoning behavior across modalities?

\textbf{RL and cross-modal transfer for multimodal models.}
Recent work has explored RL and cross-modal transfer for multimodal reasoning. Vision-R1 uses synthetic chain-of-thought data and progressive RL~\citep{huang2026visionr1incentivizingreasoningcapability}; R1-OneVision converts images into textual descriptions before SFT and RL~\citep{yang2025r1onevisionadvancinggeneralizedmultimodal}; and FAST uses difficulty-aware rewards with KL-regularized GRPO~\citep{diao2026addressingoverthinkinglargevisionlanguage}. Closest to our setting, VOLD distills from a stronger text-only LLM teacher into a VLM student using on-policy distillation and GRPO~\citep{bousselham2025voldreasoningtransferllms}. These methods show that stronger teachers can improve multimodal reasoning. In contrast, we target \emph{self-supervised} improvement: paired views of the same problem reveal cross-modal performance asymmetries that can serve as supervision for improving reasoning across modalities.

\textbf{On-policy Distillation.}
On-policy distillation has emerged as a way to learn from a teacher while reducing the distribution mismatch of standard behavioral cloning: instead of training only on teacher-generated traces, the student samples its own rollouts and receives token-level guidance from a teacher on the states it actually visits~\citep{agarwal2024onpolicydistillationlanguagemodels, gu2026minillmonpolicydistillationlarge}. More broadly, prior work on self-improvement has trained language-model agents to critique and iteratively refine their own responses through recursive self-feedback~\citep{qu2024recursive}. On-policy self-distillation provides a complementary mechanism for internal improvement, including self-distillation from demonstrations~\citep{shenfeld2026selfdistillationenablescontinuallearning}, feedback-conditioned self-teachers~\citep{hubotter2026reinforcement}, and privileged-context teachers~\citep{zhao2026selfdistilledreasoneronpolicyselfdistillation}. Recent multi-teacher or multi-model extensions~\citep{chen2026unitymas,yang2025learning,yang2025maestro} use specialized teachers, debate~\citep{yang2026auditing}, or calibrated teacher signals to merge complementary capabilities into a single student~\citep{yang2026learningteachergeneralizedonpolicy, coreteam2026mimov2flashtechnicalreport, hou2026uniopdunifyingonpolicydistillation, wang2026madopdbreakingceilingonpolicy}. \methodname{} differs in both the source and structure of the teacher signal: rather than distilling from an external stronger model, a fixed teacher pool, or privileged solution traces, it selects the strongest on-policy generations for a certain view of the same problem as a problem-specific teacher, building a self-supervised approach for improving model.

\textbf{Explicit intermediate structure for visual reasoning.}
A complementary line of work improves visual reasoning by either actively selecting informative visual evidence or introducing intermediate structures between perception and reasoning. Act2See learns to adaptively select and process visual observations for more effective video reasoning~\citep{ma2026act2see}. For visual mathematical reasoning, MathCoder-VL uses image-to-code supervision~\citep{wang2025mathcodervlbridgingvisioncode}. \citet{zhang2025openeyesreasonfinegrained} introduces a geometry-specialized vision encoder for recognizing visual primitives. Related work on plane-geometry formalization~\citep{cui2025plane} similarly exposes structured representations. These approaches introduce inductive biases through active perception, code, symbolic descriptions, formal proofs, or geometry priors. In contrast, \methodname{} retains the original input views and remains end-to-end: rather than explicitly selecting evidence or constructing intermediate representations, it encourages the model to internalize transferable reasoning structure through cross-modal supervision.

%% file: sections/pilot.tex
\section{Understanding Differences in Reasoning Across Modalities}
\label{sec:modality_diff}

Before introducing our approach, we first analyze reasoning behaviors of the base model \texttt{Qwen3-VL-4B-Instruct}~\citep{bai2025qwen3vltechnicalreport} when solving the same underlying problem from different input views. In this section, we characterize how do the model’s reasoning trajectories differ when conditioned on different modalities. We further study if access to both text and images reveal supervision signals that are hidden from any single modality?

To formalize this setting, we study post-training of a multimodal language model, denoted by $\pi_{\text{base}}$ with parameters $\theta$, on problem instances that may be presented with different input views. For a problem $\bx \sim \rho$, let $\mathcal{M}(\bx)$ denote the set of available views sharing the same underlying problem content and answer target. In this work, we focus on two views: a text-dominant view $\bx^{\text{text}}$ and an image-dominant view $\bx^{\text{img}}$. We can also readily construct a combined view $\bx^{comb} := (\bx^{\text{text}}, \bx^{\text{img}})$, which contains both views simultaneously. For any input view $\tilde{\bx} \in \mathcal{M}(\bx)$, a rollout $\by \sim \pi(\cdot \mid \tilde{\bx})$ attempts to solve the problem. As in standard outcome-reward RL, we assume that correctness can be evaluated from the final answer and define a binary reward $r(\bx,\by) \in \{0,1\}$. We study performance using pass@$k$~\citep{chen2021evaluatinglargelanguagemodels} under each modality.
\begin{table*}[t]
\centering
\caption{\footnotesize{\textbf{Representative modality-dependent reasoning behaviors.} Each column illustrates one source of cross-modal supervision; gray highlighting marks text omitted from the image-dominant query because the corresponding information is conveyed by the diagram. Full prompts and completions are in Appendix~\ref{app:examples}s.}}
\vspace{0.2cm}
\scriptsize
\setlength{\tabcolsep}{3pt}
\renewcommand{\arraystretch}{1.03}
\newcommand{\removedfromimage}[1]{{\setlength{\fboxsep}{1pt}\colorbox{gray!20}{\strut #1}}}

\begin{tabularx}{0.96\textwidth}{
    >{\raggedright\arraybackslash}p{0.09\textwidth}
    >{\raggedright\arraybackslash}p{0.25\textwidth}
    >{\raggedright\arraybackslash}p{0.25\textwidth}
    >{\raggedright\arraybackslash}p{0.32\textwidth}
}
\toprule
&
\multicolumn{1}{c}{\textbf{Example 1:} Image $\rightarrow$ Text}
&
\multicolumn{1}{c}{\textbf{Example 2:} Text $\rightarrow$ Image}
&
\multicolumn{1}{c}{\textbf{Example 3:} Image+Text $\rightarrow$ Image}
\\
\midrule

\textbf{Image}
&
\multicolumn{1}{c}{\includegraphics[width=0.16\textwidth]{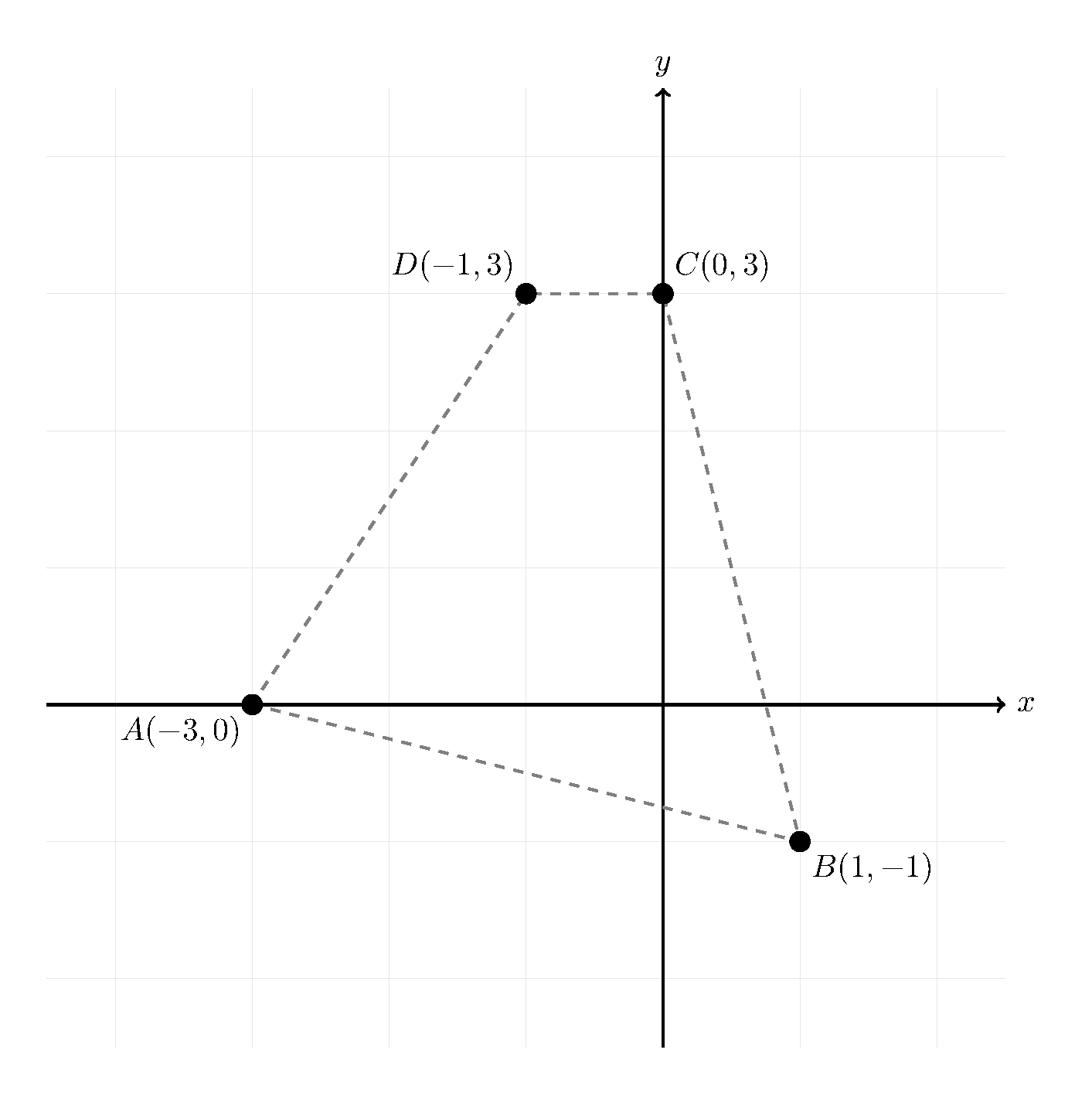}}
&
\multicolumn{1}{c}{\includegraphics[width=0.18\textwidth]{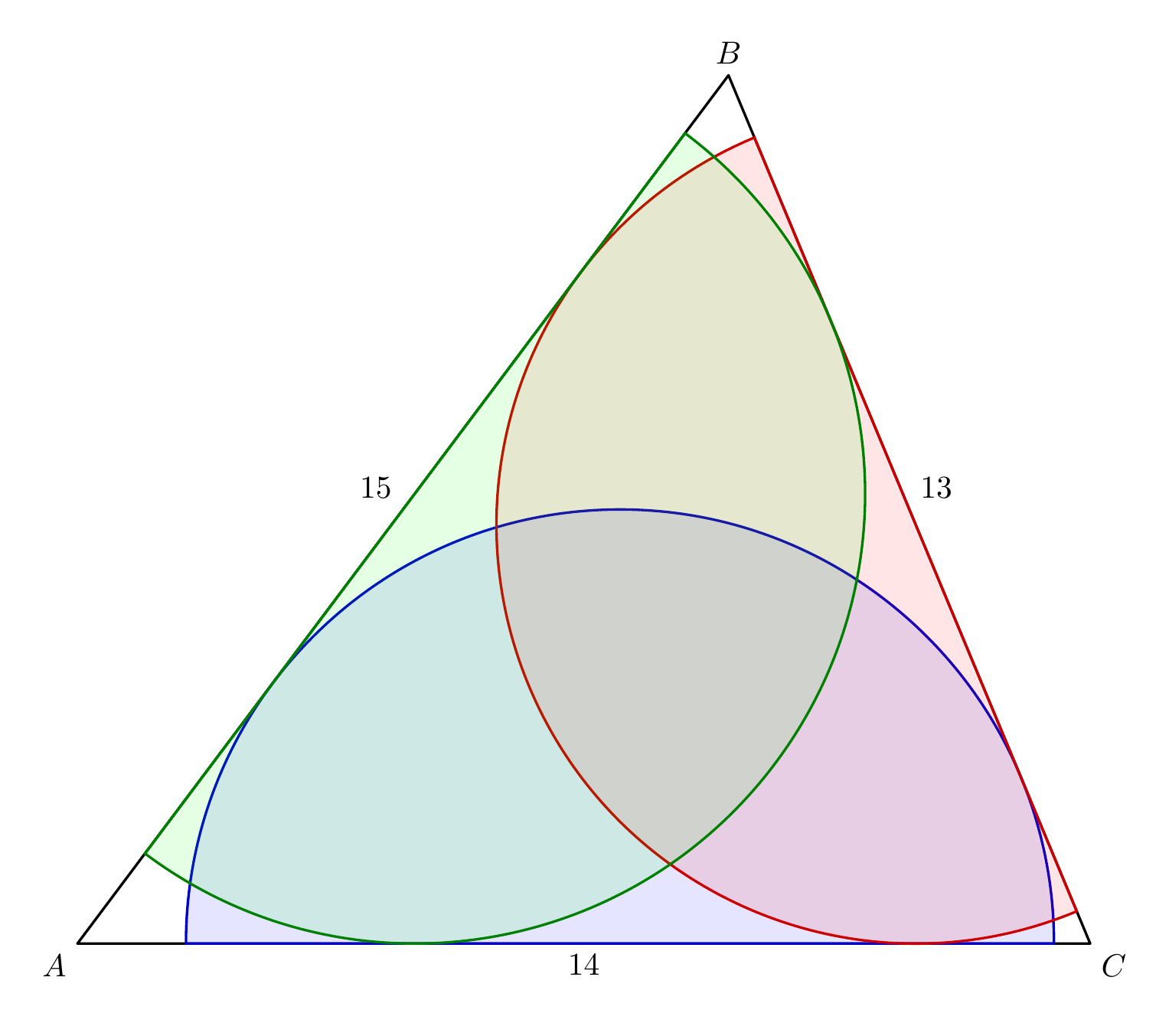}}
&
\multicolumn{1}{c}{\includegraphics[width=0.12\textwidth]{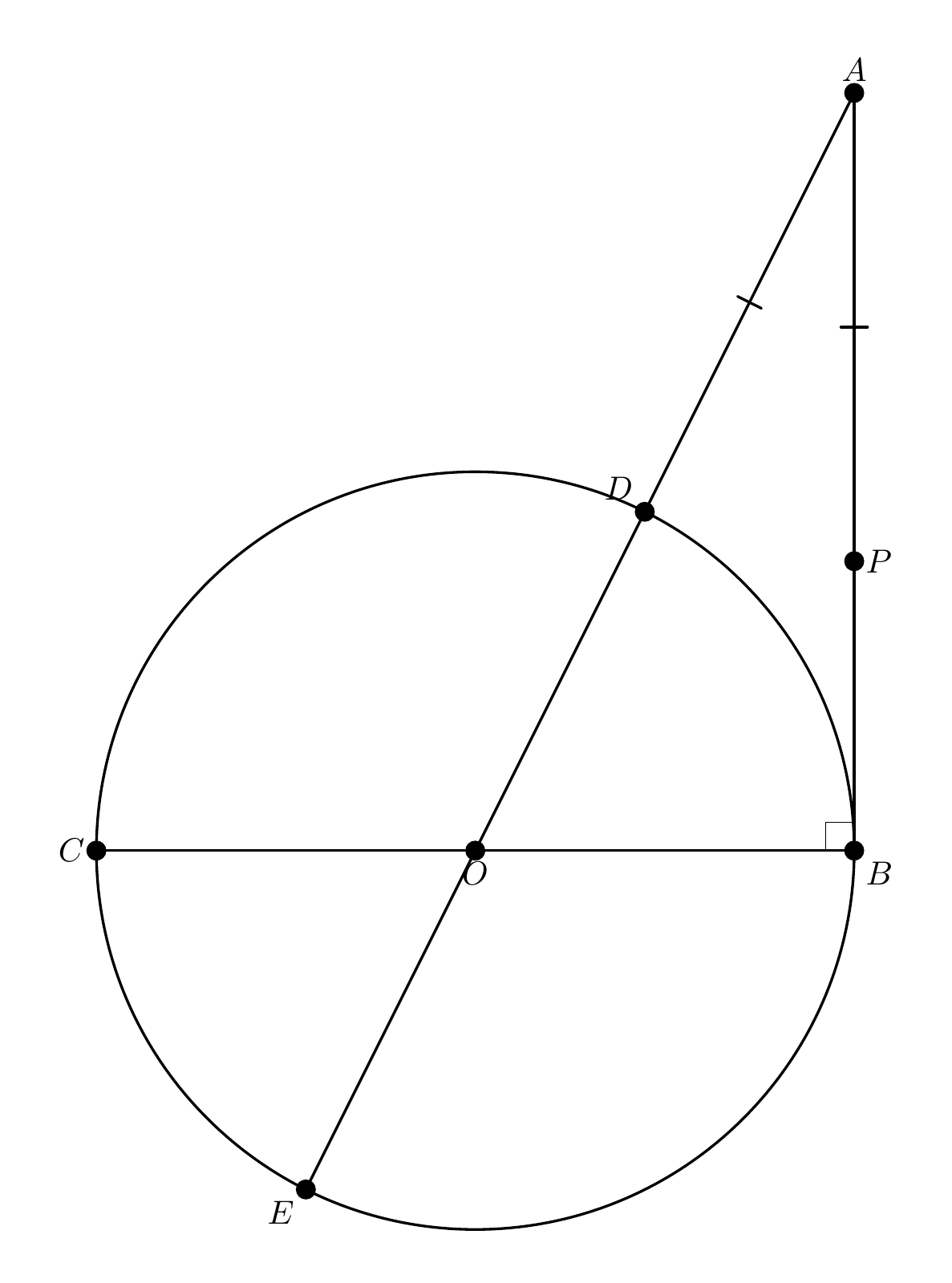}}
\\

\midrule
\textbf{Query}
&
Given four fixed points $A$\removedfromimage{$(-3,0)$}, $B$\removedfromimage{$(1,-1)$}, $C$\removedfromimage{$(0,3)$}, $D$\removedfromimage{$(-1,3)$} and a moving point $P$, find
$\min_P |PA|+|PB|+|PC|+|PD|$.
&
In a \removedfromimage{$13$-$14$-$15$} triangle, three semicircles are drawn with diameters on the sides and tangent to the other two sides. The product of their areas equals the cube of a sphere's surface area. If the sphere volume is $\frac{a\pi\sqrt b}{c}$, find $a+b+c$.
&
In circle $O$, \removedfromimage{$BC$ is a diameter,} \removedfromimage{$AB \perp BC$,} \removedfromimage{$ADOE$ is a line,} \removedfromimage{$AP=AD$,} and $AB=2R$. Which option is correct?
(A) $AP^2=PB \cdot AB$.
(B) $AP \cdot DO=PB \cdot AD$.
(C) $AB^2=AD \cdot DE$.
(D) $AB \cdot AD=OB \cdot AO$.
(E) None.
\\

\midrule
\textbf{Before training}
&
\textbf{Image-dominant:} solves correctly with $3\sqrt{2}+2\sqrt{5}$.

\textbf{Text-dominant:} performs exhaustive search of possible locations of $P$ and gives a rough estimate $8.74$. Actually, $P$ at the intersection of $AC$ and $BD$ minimizes the desired quantity.
&
\textbf{Text-dominant:} solves correctly with $873$.

\textbf{Image-dominant:} misreads each semicircle's diameter as exactly the length of each side of the triangle. The model wrongly predicts $3249$. Actually, each diameter is shorter than the side.
&
\textbf{Combined image+text:} uses the textual relation that explicitly stated $AP=AD$ and correctly chooses A.

\textbf{Image-dominant:} misses $AP=AD$ condition from the diagram and wrongly chooses B.

\\

\midrule
\textbf{After training}
&
\textbf{Text-dominant:} recovers the structural solution and predicts $3\sqrt{2}+2\sqrt{5}$.
&
\textbf{Image-dominant:} avoids the visual grounding error and predicts $873$.
&
\textbf{Image-dominant:} notes the tick marks indicating $AD=AP$ and now chooses A.
\\

\midrule
\textbf{Takeaway}
&
The diagram makes the geometric structure salient, so image-dominant reasoning supervises text-dominant reasoning.
&
The text removes a perceptual bottleneck, so text-dominant reasoning supervises image-dominant reasoning.
&
Combined image+text view leverages explicit textual descriptions together with the visual structure, removing ambiguity. So it supervises image-dominant reasoning.
\\

\bottomrule
\end{tabularx}

\label{tab:modality_examples_compact}
\end{table*}
\label{sec:data_introduction}
To study view-dependent reasoning, we construct \textbf{ODA-Data}, a geometry dataset derived from ODA-Math-460k~\citep{cai2025opendataarena}. Each example contains two views of the same problem: a ``text-dominant'' view and an ``image-dominant'' view. In our analysis and training, we also consider a third combined image+text view, which is constructed on the fly by concatenating the text-dominant prompt with the image from the image-dominant view. Starting from the pure text dataset ODA-Math-460k, we first exclude easy problems whose original-dataset annotations report a pass rate of 1. For each remaining problem, we use Gemini-3-Pro-Preview~\citep{google2025gemini3pro} to generate and verify a diagram in TikZ, a LaTeX-based language for programmatic scientific graphics, and rewrite the image-dominant prompt so that it no longer contains information present in the diagram. We choose TikZ over direct image generation because preliminary human inspection indicated that the model generated usable TikZ code more reliably and that the compiled diagrams had higher visual quality than directly generated images. In this process, diagrams that do not accurately reflect the question or contain answer leakage are removed. To construct a challenging dataset in which every problem is difficult under at least one view, we generate 32 rollouts per view with Qwen3-VL-4B-Instruct~\citep{bai2025qwen3vltechnicalreport} and retain problems for which exactly one view has at most one successful rollout, yielding approximately 2K curated examples. See Appendix~\ref{sec:prompting} for full details. The dataset is randomly split 85:15 into \textbf{ODA-Train} for post-training and \textbf{ODA-Val} for evaluation. Because ODA-Data is filtered using this view-asymmetric failure criterion, ODA-Val serves as a diagnostic benchmark for studying cross-view performance differences and transfer.

We compare reasoning traces induced by the text-dominant, image-dominant, and combined views. Although these views share the same problem and answer target, they expose different information and can induce different reasoning trajectories. Text-conditioned reasoning can use explicitly stated relations directly, whereas image-conditioned reasoning must first recover them from the diagram, creating a potential grounding bottleneck. For example, in the $13$-$14$-$15$ triangle problem in Table~\ref{tab:modality_examples_compact}, the text-conditioned trace correctly derives $873$, whereas the image-conditioned trace incorrectly treats each semicircle's diameter as an entire triangle side and predicts $3249$. Conversely, diagrams can make geometric structure salient when text-only reasoning fails, while the combined view can help when neither restricted view is fully reliable. These complementary strengths motivate adaptive teacher selection across the three views.

\textbf{Quantitative analysis.} We next quantify these view-dependent differences over ODA-Train using \texttt{Qwen3-VL-4B-Instruct} and examine pass@$1$ metric. At the aggregate level, text-dominant achieves 29.58\%, image-dominant achieves 10.54\%, and combined image+text prompting achieves 26.84\%. Across ODA-Train, the text-dominant view performs best on 51.14\% of problems, the image-dominant view on 37.98\%, and the combined image+text view on 10.88\%. Thus, although text-dominant prompting is strongest on average, nearly half of the problems are best supervised by a non-text-dominant view. These results suggest that the qualitative behaviors in Table~\ref{tab:modality_examples_compact} are not isolated anecdotes, but recurring sources of modality asymmetry.

Given these asymmetries, a natural question is whether standard RL can exploit multiple views simply by training on them together. A controlled single-problem experiment in Appendix~\ref{app:single_problem_analysis} shows that naive mixed-modality GRPO underperforms GRPO trained on the corresponding single view, indicating that exposure to paired views alone does not reliably transfer successful reasoning across modalities. We next introduce \methodname{}, which makes this transfer explicit by converting the strongest view of each problem into a teacher signal for training the student views.

%% file: sections/methodology.tex
\section{Our Approach: \methodname{}}
\label{sec:method}

Our goal is to improve reasoning under a student view, either the text-dominant view or the image-dominant view. Section~\ref{sec:modality_diff} shows that no single view is uniformly best, and \methodname{} converts this view-dependent heterogeneity into a self-supervision signal. For each problem, it selects the view on which the current model performs the best as a problem-specific teacher. The restricted-view student is trained with standard outcome-reward RL plus an auxiliary  reverse-KL regularizer. Crucially, the student always generates rollouts from its own restricted view; the selected teacher view is used only to score those student-generated trajectories.

\textbf{Best-modality teacher selection.}
Each problem has three candidate teacher views: text-dominant $\bx^{\text{text}}$, image-dominant $\bx^{\text{img}}$, and combined image+text $\bx^{\text{comb}}=(\bx^{\text{text}},\bx^{\text{img}})$. Let $\mathcal{M}=\{\text{text},\text{img},\text{comb}\}$ denote the candidate teacher views. 
Teacher selection is performed online during training using the current policy model. 
For each problem $\bx$ and each candidate view $j\in\mathcal{M}$, we sample $K=16$ rollouts and estimate the view-specific success rate. We then select the best-performing view, breaking ties uniformly at random. The teacher distribution is then defined as a slow-moving copy of the model conditioned on the selected teacher view, $\pi_{\text{teach}}(\cdot\mid\bx)
=
\pi_{\bar\theta}(\cdot\mid \bx^{(j^\star(\bx))})$, where $\bar\theta$ is an exponential moving average (EMA) of the policy parameters, described below.
Thus, the teacher direction is selected independently for each problem and can adapt throughout training, rather than being fixed globally or precomputed from the base model. Although teacher selection considers three candidate views, the text-dominant and image-dominant rollouts are already generated as standard online GRPO rollouts; thus, the only additional rollout generation required by \methodname{} is for the combined image+text view, which introduces negligible computational overhead. We provide a FLOPs comparison with mixed-modality GRPO in Appendix~\ref{app:compute}.

\textbf{Training under a restricted view.}
For a restricted view $m\in\{\text{text},\text{img}\}$, the student receives the corresponding input $\bx^{(m)}$ and samples rollouts
$\by\sim\pi_\theta(\cdot\mid\bx^{(m)})$. 
The selected teacher may condition on a different view $\bx^{(j^\star(\bx))}$, including the combined image+text view. The challenge is therefore to use the stronger teacher view without training the student on off-policy teacher-generated trajectories or forcing it to rely on information unavailable from its restricted input. We address this by optimizing a weighted sum of outcome-reward RL and an on-policy reverse-KL regularizer:
\begin{align}
\label{eq:full_loss}
\mathcal{L}(\theta;m)
=
\mathcal{L}_{\text{RL}}(\theta;m)
+
\lambda_{\text{KL}}\mathcal{L}_{\text{rKL}}(\theta;m)
\end{align}
where $\lambda_{\text{KL}}\ge 0$ controls the strength of teacher guidance. 
The RL term is a GRPO-style on-policy objective that optimizes final-answer correctness under the restricted student view. 
The reverse-KL term, defined below, transfers token-level preferences from the selected teacher view while keeping rollouts on-policy with respect to the restricted-view student.

\textbf{Reverse KL for cross-view distillation.}
The auxiliary loss distills the selected teacher's reasoning preferences while keeping rollouts on-policy: for a student view $m\in\{\text{text},\text{img}\}$, we sample $\by\sim\pi_\theta(\cdot\mid\bx^{(m)})$ and use the selected teacher only to rescore the same generated tokens. This gives the reverse-KL value
\begin{equation}
\label{eq:rkl_obj}
\providecommand{\mirrorRKLEquationWidth}{\dimexpr\linewidth-3.5em\relax}
\resizebox{\mirrorRKLEquationWidth}{!}{$\displaystyle
\mathcal{L}_{\text{rKL}}(\theta;m)
=
\mathbb{E}_{\bx,\by \sim \pi_\theta(\cdot \mid \bx^{(m)})}
\left[
\sum_{t=1}^{|\by|-1}
\log
\frac{\pi_\theta(y_{t+1}\mid \bx^{(m)},y_{1:t})}
{\pi_{\bar\theta}(y_{t+1}\mid \bx^{(j^\star(\bx))},y_{1:t})}
\right]
$}
\end{equation}
To avoid an expensive full-vocabulary KL computation, we instead optimize the sampled K3 surrogate $g(r)=e^{-r}+r-1$, where $r=\log\pi_\theta(y_{t+1}\mid\cdot)-\log\pi_{\bar\theta}(y_{t+1}\mid\cdot)$ and $\mathbb{E}_{\pi_\theta}[g(r)]=D_{\mathrm{KL}}(\pi_\theta\|\pi_{\bar\theta})$. Operationally, this asks how compatible the student's tokens are with the stronger teacher view only at states induced by the policy being trained, avoiding unnecessary distribution shift.

\textbf{EMA teacher.}
Because the teacher in Equation~\ref{eq:rkl_obj} is derived from the same network being trained, the most direct choice is to score the teacher term with the current parameters $\theta$. However, this makes the distillation target move every step: the student chases a target that its own updates displace. Empirically, this self-referential coupling induces a slow inflation of policy entropy and reference-policy KL that eventually destabilizes long-horizon training (Figure~\ref{fig:ema_teacher_stability}). We therefore parameterize the teacher with an exponential moving average (Polyak average) of the policy parameters,
\begin{align}
\label{eq:ema_update}
\bar\theta \;\leftarrow\; \alpha\,\bar\theta + (1-\alpha)\,\theta
\end{align}
applied after every policy update, with $\bar\theta$ initialized to $\theta$ at the beginning of training. This mirrors target networks in value-based RL~\citep{mnih2015human,lillicrap2019continuouscontroldeepreinforcement} and weight-averaged teachers in self-supervised learning~\citep{tarvainen2017mean,he2020momentum,grill2020bootstrap}: the teacher tracks the improving policy on a slower timescale, providing a slowly improving but stable target. The EMA teacher is held fixed during each update, while its token probabilities determine the direction and magnitude of the student gradients through the K3 surrogate.

The two terms in Equation~\ref{eq:full_loss} play complementary roles: the GRPO term optimizes final-answer correctness under the restricted view, while the reverse-KL term provides dense token-level guidance from the strongest view on the student's own trajectories. 
Unlike methods that sample from or imitate an external teacher distribution, \methodname{} keeps rollouts on-policy with respect to the restricted-view student and uses the selected teacher view only as a supervision signal. Please see Algorithm~\ref{alg:mirror} for the full pseudocode of \methodname{} update.

%% file: sections/experiment.tex
\section{Experimental Evaluation}
\label{sec:exps}

\definecolor{gainSbg}{HTML}{F1F6ED}
\definecolor{gainMbg}{HTML}{E7F1DF}
\definecolor{gainLbg}{HTML}{DDECCF}

\definecolor{lossSbg}{HTML}{FCF1F1}
\definecolor{lossMbg}{HTML}{F9E5E5}
\definecolor{lossLbg}{HTML}{F5D8D8}

\newcommand{\deltatag}[2]{%
  \nobreak\hspace{0.18em}%
  \raisebox{0.10ex}{%
    \tikz[baseline=(delta.base)]{%
      \node[
        fill=#1,
        draw=none,
        rounded corners=1.4pt,
        inner xsep=1.15pt,
        inner ysep=0.25pt,
        text height=1.35ex,
        text depth=0.20ex
      ] (delta) {\fontsize{5.5}{6.0}\selectfont #2};%
    }%
  }%
}

\newcommand{\gainS}[1]{\deltatag{gainSbg}{#1}}
\newcommand{\gainM}[1]{\deltatag{gainMbg}{#1}}
\newcommand{\gainL}[1]{\deltatag{gainLbg}{#1}}

\newcommand{\lossS}[1]{\deltatag{lossSbg}{#1}}
\newcommand{\lossM}[1]{\deltatag{lossMbg}{#1}}
\newcommand{\lossL}[1]{\deltatag{lossLbg}{#1}}

\newcommand{\grp}[1]{\multicolumn{1}{l}{\emph{#1}}}
\newcommand{\ind}[1]{\hspace{1.2em}#1}

The goal of our experiments is to evaluate whether \methodname{} can turn asymmetries across views into an effective self-supervision signal for improving multimodal reasoning. We answer four questions in this section:
\textbf{(1)} Does MIRROR improve restricted-view reasoning over single-view and mixed-modality GRPO baselines? \textbf{(2)} Does selecting the teacher view separately for each problem outperform using a fixed global teacher view?
\textbf{(3)} Does \methodname{} expand the set of solvable problem instances after training?
\textbf{(4)} How sensitive is \methodname{} to the cross-view KL coefficient $\lambda_\text{KL}$ and to the parameterization of the teacher (current policy vs.\ slow-moving EMA)?

\textbf{Experimental setup.}
We post-train \texttt{Qwen3-VL-4B-Instruct} and implement \methodname{} in verl~\citep{sheng2025hybridflow}. Models are trained with maximum response length $16{,}384$, sampling temperature $0.8$, GRPO clipping range $\epsilon_{\mathrm{low}}=0.2,\epsilon_{\mathrm{high}}=0.26$, reference-policy KL coefficient $0.001$, and entropy coefficient $0.001$. \methodname{} uses a reverse-KL coefficient of 0.01 and EMA weights $\alpha=0.99$. Our in-distribution evaluation uses ODA-Val, where paired views preserve problem identity and allow us to measure pass@$k$ under each view and track how training changes problem-level solvability across views. To test whether improvements extend beyond this asymmetry-enriched distribution, we also evaluate on a disjoint random subset of unfiltered ODA-Math-460K that excludes all samples used in ODA-Train and ODA-Val. We additionally evaluate on public multimodal geometry benchmarks GeoInt~\citep{wei2025geoint} and MathVerse~\citep{zhang2024mathversedoesmultimodalllm}. For GeoInt, we consider only answer-based question. We evaluate final-answer correctness with pass@1 and pass@16 estimated by 32 rollouts. For MathVerse, we use \texttt{gpt-4o-2024-11-20}~\citep{hurst2024gpt} as the judge through the LMMs-Eval suite~\citep{zhang2024lmmsevalrealitycheckevaluation} and report the mean score.

\begin{table}[t]
\centering
\caption{\footnotesize
\textbf{\methodname{} improves single-view reasoning over standard GRPO.}
We report pass@1 and pass@16 on held-out image/text ODA-Val
problems and GeoInt, pass@1 on a disjoint random unfiltered ODA-460K subset,
and mean judge score on MathVerse.
Parenthetical values after model or method names denote the reported number of post-training samples.
Green tags indicate improvement and pink rounded tags indicate
degradation relative to the Qwen3-VL-4B Instruct base model.
\textbf{Bold} numbers indicate the best result.
}
\vspace{0.2cm}

\resizebox{\linewidth}{!}{%
\begin{tabular}{@{}l c c c c c c c c c@{}}
\toprule

\textbf{Approach}
&
\multicolumn{2}{c}{\textbf{ODA-Val Image}}
&
\multicolumn{2}{c}{\textbf{ODA-Val Text}}
&
\multicolumn{2}{c}{\textbf{ODA-Unfiltered}}
&
\multicolumn{2}{c}{\textbf{GeoInt}}
&
\textbf{MathVerse}
\\

\cmidrule(lr){2-3}
\cmidrule(lr){4-5}
\cmidrule(lr){6-7}
\cmidrule(lr){8-9}
\cmidrule(lr){10-10}

&
pass@1
&
pass@16
&
pass@1
&
pass@16
&
\shortstack{image\\pass@1}
&
\shortstack{text\\pass@1}
&
pass@1
&
pass@16
&
mean
\\

\midrule

\grp{Existing models}
& & & & & & & & &
\\

\ind{Vision-R1-7B (210K)}
& 7.41
& 19.48
& 9.67
& 29.35
& 11.40
& 11.10
& 23.30
& 52.80
& 46.75
\\

\ind{PAPO-7B (39K)}
& 11.38
& 25.77
& 16.88
& 39.57
& 17.80
& 17.90
& 29.90
& 60.30
& 47.82
\\

\ind{Vero-8B (600K)}
& 16.16
& 39.74
& 35.48
& 71.73
& 34.50
& 38.90
& \textbf{67.30}
& 77.42
& \textbf{69.49}
\\

\midrule

\grp{Base model}
& & & & & & & & &
\\

\ind{Qwen3-VL-4B Instruct}
& 12.30
& 42.57
& 32.91
& 80.22
& 35.40
& 35.89
& 58.62
& 70.79
& 41.22
\\

\midrule

\grp{Standard GRPO (2K)}
& & & & & & & & &
\\

\ind{text-dominant GRPO}
& 17.32 \gainM{+5.02}
& 43.29 \gainS{+0.72}
& 39.92 \gainM{+7.01}
& 81.06 \gainS{+0.84}
& 36.40 \gainS{+1.00}
& 37.40 \gainS{+1.51}
& 63.02 \gainM{+4.40}
& 78.95 \gainL{+8.16}
& 45.38 \gainM{+4.16}
\\

\ind{image-dominant GRPO}
& 20.38 \gainL{+8.08}
& 48.78 \gainM{+6.21}
& 41.06 \gainL{+8.15}
& 83.16 \gainM{+2.94}
& 39.10 \gainM{+3.70}
& 39.80 \gainM{+3.91}
& 62.30 \gainM{+3.68}
& \textbf{79.69} \gainL{+8.90}
& 39.24 \lossS{$-$1.98}
\\

\ind{mixed-modality GRPO}
& 17.99 \gainM{+5.69}
& 45.68 \gainM{+3.11}
& 39.18 \gainM{+6.27}
& 81.66 \gainS{+1.44}
& 36.87 \gainS{+1.47}
& 39.15 \gainM{+3.26}
& 61.68 \gainS{+3.06}
& 75.21 \gainM{+4.42}
& 44.72 \gainS{+3.50}
\\

\midrule

\ind{\methodname{} (2K)}
& \textbf{23.57} \gainL{+11.27}
& \textbf{57.06} \gainL{+14.49}
& \textbf{45.67} \gainL{+12.76}
& \textbf{86.10} \gainM{+5.88}
& \textbf{39.40} \gainM{+4.00}
& \textbf{41.29} \gainM{+5.40}
& 66.15 \gainL{+7.53}
& 78.38 \gainL{+7.59}
& 46.78 \gainM{+5.56}
\\

\bottomrule
\end{tabular}%
}
\label{tab:main_results}

\end{table}

\textbf{Baselines and comparisons.}
We evaluate the base \texttt{Qwen3-VL-4B-Instruct} model and three post-trained multimodal reasoners: \texttt{Vision-R1-7B}~\citep{huang2026visionr1incentivizingreasoningcapability}, which applies cold-start training and RL to \texttt{Qwen2.5-VL-7B-Instruct}; \texttt{PAPO-7B}~\citep{wang2025perception}, which adds a perception-aware objective contrasting original and masked-image rollouts to the same base model; and \texttt{Vero-8B}~\citep{sarch2026veroopenrlrecipe}, which applies a broad, task-routed visual-reasoning recipe to \texttt{Qwen3-VL-8B-Instruct}.
Our controlled ODA-Train comparisons include text-dominant and image-dominant single-view GRPO, mixed-modality GRPO trained on both views, and fixed-teacher \methodname{} variants. Mixed-modality GRPO tests whether cross-view transfer emerges implicitly when both views are present, while \methodname{} uses the same mixture of student prompts but makes the transfer direction explicit. We provide additional results on other base models in Appendix~\ref{app:qwen8b}. We also show \methodname{}'s generalization to the chart domain in Appendix~\ref{app:chart_generalization}.

\textcolor{lightblue}{\textbf{Result 1: \methodname{} improves over standard and mixed-modality GRPO.}}
Table~\ref{tab:main_results} compares \methodname{} with the base model, existing reasoning-based VLMs, and standard RL baselines. On ODA-Val, \methodname{} performs best across all image- and text-view metrics, outperforming both the strongest single-view GRPO baseline and mixed-modality GRPO.
On the ODA-Unfiltered, the base model performs nearly identically across the two views, confirming that this is not an asymmetry enriched distribution. Yet \methodname{} still outperforms all baselines compared. Thus, the benefit of cross-view supervision is not limited to the filtered distribution that exposes strong modality asymmetry.
The gains also transfer beyond ODA-Val: \methodname{} improves over the base model on GeoInt and MathVerse. Despite using only 2K training examples, it outperforms Vision-R1-7B and PAPO-7B across ODA-Val and GeoInt and Vero-8B across ODA-Val, although Vero-8B remains strongest on MathVerse. Appendix~\ref{app:qwen8b} further shows that the gains generalize to Qwen3-VL-8B-Instruct~\citep{bai2025qwen3vltechnicalreport} and GLM-4.1V-9B-Thinking~\citep{glm-vteam2025glm41vthinking}, with \methodname{} outperforming standard GRPO across all reported metrics for both larger base models. Together, these results show that cross-view asymmetry provides a useful training signal across evaluations, model scales, and model families.

\begin{wrapfigure}{r}{0.49\textwidth}
\centering

\includegraphics[width=0.96\linewidth]{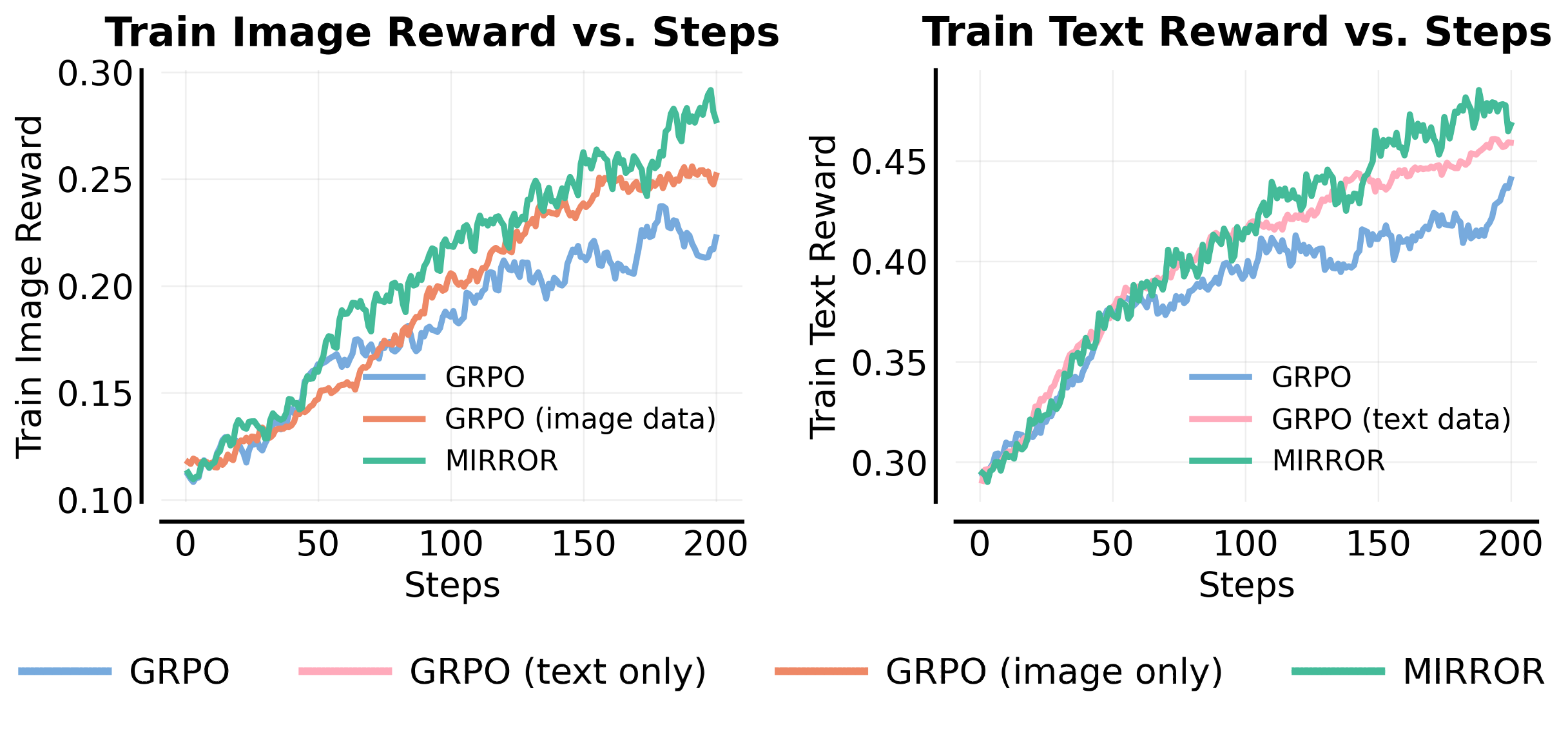}
\vspace{-0.2cm}
\caption{\footnotesize
\textbf{Mixed-modality GRPO does not automatically transfer reasoning across views.}
With full-dataset training, mixed-view GRPO underperforms GRPO trained directly on the corresponding restricted view, while \methodname{} obtains the strongest training reward by using paired views directionally.}
\label{fig:mixed_modality_transfer}
\vspace{-0.2cm}

\end{wrapfigure}

Mixed-modality GRPO is the most direct test of whether paired views are sufficient by themselves. Although this baseline trains on both text-dominant and image-dominant views of the same underlying problems, it treats each view as an independent RL prompt and relies on transfer emerging implicitly through shared parameters. Figure~\ref{fig:mixed_modality_transfer} and ~\ref{fig:single_problem_analysis}  show that this is not reliable: mixed-view GRPO obtains lower training reward than single-view GRPO in both the controlled setting and full-dataset training. This behavior is also consistent with \emph{ray interference} in heterogeneous on-policy RL mixtures~\citep{qu2026popelearningreasonhard}. In our setting, views differ in grounding requirements, and access to explicit relations. Optimizing all views with sparse outcome rewards does not guarantee that updates from an easier view teach the harder restricted view. \methodname{} addresses this by specifying the direction of transfer: it selects the strongest-performing view to regularize the restricted-view student toward its teacher distribution. As detailed in Appendix~\ref{app:compute}, \methodname{} incurs approximately $37.5\%$ more FLOPs per update than standard mixed-modality GRPO. To control for this difference, we compare \methodname{} at step 150 with GRPO at step 200, which have roughly matched cumulative compute. Even under this compute-matched comparison, \methodname{} achieves higher image ($0.2625$ vs.\ $0.2368$) and text ($0.4581$ vs.\ $0.4387$) reward, suggesting that its gains cannot be attributed to additional computation alone.

\begin{table}[t]
\centering
\caption{\footnotesize \textbf{Ablation on teacher choice.}
We compare fixed teacher choices against \methodname{}, which adaptively selects the strongest modality-specific teacher for each problem.
By selecting the teacher per problem, \methodname{} better exploits modality-dependent strengths and provides more consistent cross-modal guidance.}
\vspace{0.2cm}
\resizebox{\linewidth}{!}{%
\begin{tabular}{@{}l c c c c c c c@{}}
\toprule
\textbf{Teacher choice} &
\multicolumn{2}{c}{\textbf{ODA-Val Image}} &
\multicolumn{2}{c}{\textbf{ODA-Val Text}} &
\multicolumn{2}{c}{\textbf{GeoInt}} &
\textbf{MathVerse} \\
\cmidrule(lr){2-3}\cmidrule(lr){4-5}\cmidrule(lr){6-7}\cmidrule(lr){8-8}
& pass@1 & pass@16 & pass@1 & pass@16 & pass@1 & pass@16 & mean \\
\midrule
\ind{text-dominant teacher}
& 19.12 & 51.03 & 39.78 & 83.59 & 64.41 & \textbf{79.59} & 45.28 \\
\ind{image-dominant teacher}
& 18.95 & 49.10 & 43.43 & 83.77 & 63.80 & 79.50 & 43.36 \\
\ind{combined image+text teacher}
& 20.99 & 52.28 & 43.35 & 82.33 & 64.56 & 78.77 & \textbf{47.26} \\
\ind{Adaptive teacher (\methodname{})}
& \textbf{23.57} & \textbf{57.06}
& \textbf{45.67} & \textbf{86.10}
& \textbf{66.15} & 78.38
& 46.78 \\
\bottomrule
\end{tabular}%
}

\label{tab:teacher_choice}

\end{table}

\textcolor{lightblue}{\textbf{Result 2: Adaptive teacher selection provides the most consistent gains.}}
We next evaluate whether the teacher view should be fixed globally.
Table~\ref{tab:teacher_choice} shows that the fixed-teacher variants have complementary strengths, but none is uniformly best. These differing preferences indicate that the most useful supervision source is problem-dependent rather than governed by one global transfer direction.
By selecting the strongest view separately for each problem, \methodname{} achieves the best result on most metrics. Appendix~\ref{app:teacher_dynamics} further shows that teacher-selection proportions evolve during training as the policy's view-specific capabilities change. Together, these results motivate selecting teachers online and per problem rather than fixing them globally or from the base model.

\begin{wrapfigure}{r}{0.49\textwidth}
\centering
\vspace{-0.1cm}
\includegraphics[width=0.93\linewidth]{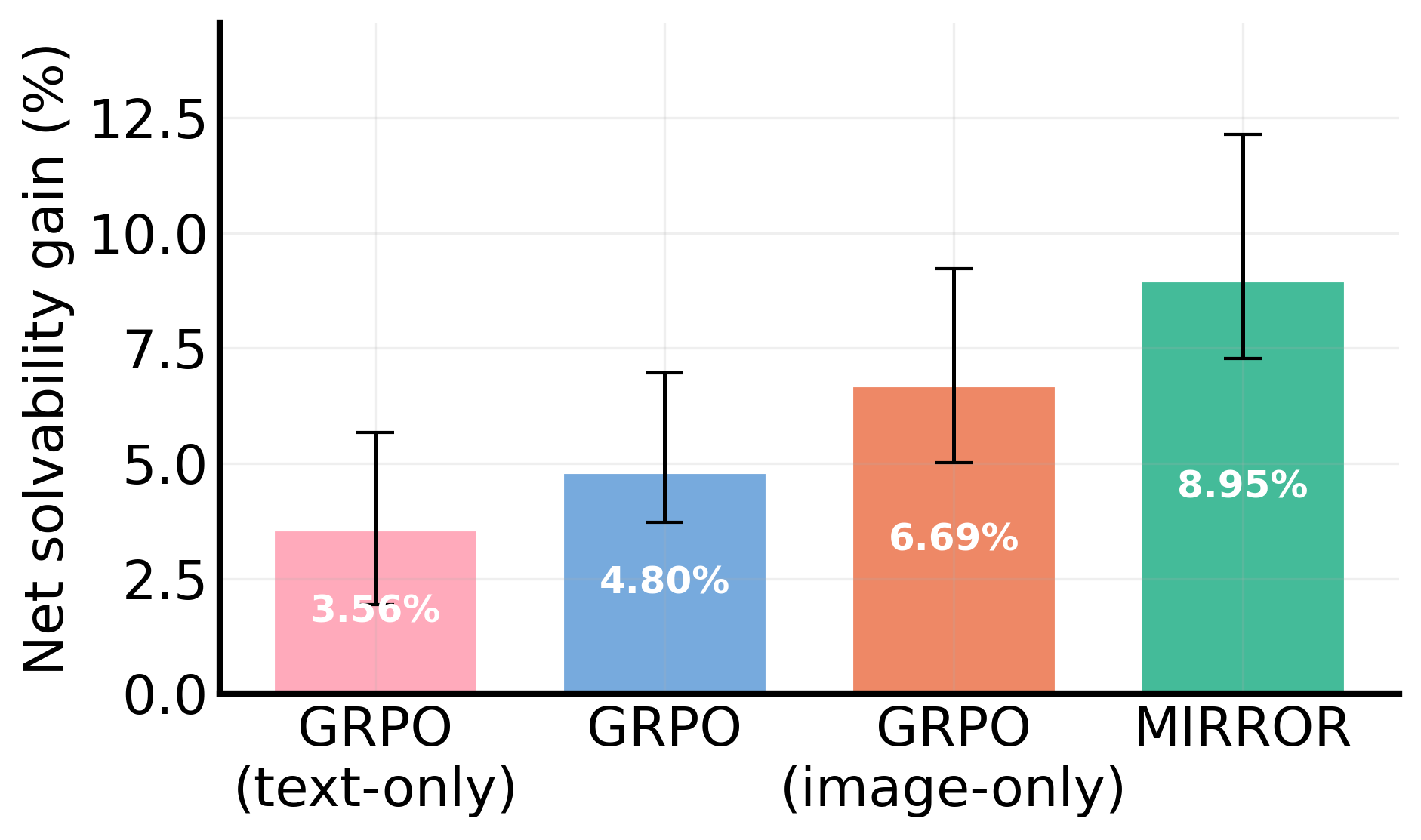}
\vspace{-0.2cm}
\caption{\footnotesize
\textbf{Net solvability gain after training.}
Each bar shows the change in the fraction of solved examples relative to the base model across 3 random seeds. Each random seed samples 309 problems which is the same size as the validation set. Error bars indicate the min and max values across seeds. \methodname{} produces the largest net increase.}
\vspace{-0.4cm}
\label{fig:solvability_shift}
\end{wrapfigure}
\textcolor{lightblue}{\textbf{Result 3: \methodname{} expands solvability and improves cross-view consistency.}}
Aggregate pass@$k$ scores do not fully capture how post-training changes the set of problems a model can solve.
We therefore measure solvability on ODA-Train before and after training.
Net solvability gain is the before-to-after change in the percentage of sampled problems with at least one successful rollout. As shown in Figure~\ref{fig:solvability_shift}, all GRPO variants improve net solvability, but \methodname{} achieves the largest gain at $+8.95$ pp. Beyond expanding the set of solvable examples, \methodname{} also improves cross-view consistency.
On ODA-Train, only $42.5\%$ of problems are solvable under both the text-dominant and image-dominant views by the base model.
Standard GRPO improves this both-view solvability rate to $53.6\%$, while \methodname{} further increases it to $60.7\%$.
This shows that \methodname{} not only improves accuracy under each restricted view, but also helps the model solve the same underlying problem consistently across modalities.

\begin{table}[t]
\centering
\caption{\footnotesize \textbf{Ablation on the reverse-KL coefficient.}
We sweep the strength of the cross-modal reverse-KL regularizer used in \methodname{} while keeping all other components fixed and let it ran for 100 steps.
This ablation tests how sensitive \methodname{} is to the alignment strength and identifies the regime that best balances on-policy RL with cross-modal distillation.}
\vspace{0.2cm}
\resizebox{\linewidth}{!}{%
\begin{tabular}{@{}l c c c c c c c@{}}
\toprule
\textbf{Reverse KL coefficient} &
\multicolumn{2}{c}{\textbf{ODA-Val Image}} &
\multicolumn{2}{c}{\textbf{ODA-Val Text}} &
\multicolumn{2}{c}{\textbf{GeoInt}} &
\textbf{MathVerse} \\
\cmidrule(lr){2-3}\cmidrule(lr){4-5}\cmidrule(lr){6-7}\cmidrule(lr){8-8}
& pass@1 & pass@16 & pass@1 & pass@16 & pass@1 & pass@16 & mean \\
\midrule
$\lambda_{\mathrm{KL}} = 0.001$
& 18.26
& 49.98
& 42.03
& 85.21
& 62.30
& 77.24
& 44.50 \\
$\lambda_{\mathrm{KL}} = 0.005$
& 17.74
& 49.45
& 40.83
& \textbf{86.70}
& 62.13
& 76.71
& 45.41 \\
$\lambda_{\mathrm{KL}} = 0.01$
& \textbf{21.74}
& {52.36}
& \textbf{44.44}
& {85.25}
& \textbf{64.58}
& \textbf{79.70}
& \textbf{46.98} \\
$\lambda_{\mathrm{KL}} = 0.05$
& 18.75
& \textbf{54.73}
& 39.67
& 85.07
& 61.46
& 77.17
& 46.62 \\
\bottomrule
\end{tabular}%
}

\label{tab:rankkl_coef}
\vspace{-0.3cm}
\end{table}

\textcolor{lightblue}{\textbf{Ablation 1: effect of the reverse-KL
coefficient.}}
Table~\ref{tab:rankkl_coef} studies the strength of the reverse-KL regularizer while keeping all other training components fixed for 100 steps. We found that $\lambda_{\text{KL}}=0.01$ achieves the best rank overall and therefore use it for subsequent training. Larger coefficients degraded performance, indicating that excessive alignment can overpower the RL objective.

\begin{figure}[t]
\centering
\includegraphics[width=\linewidth]{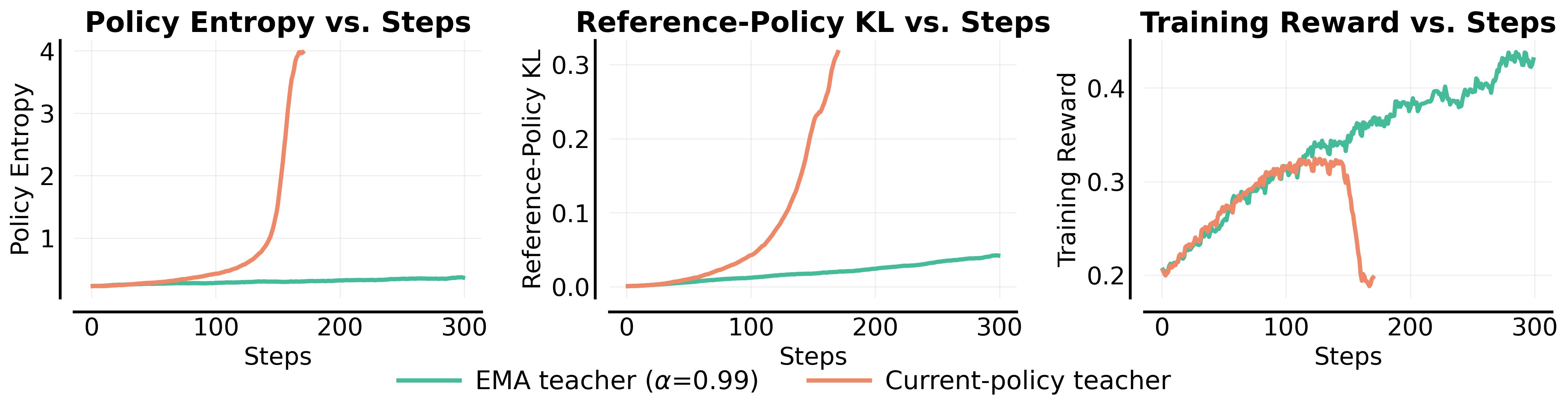}
\vspace{-0.8cm}
\caption{\footnotesize
\textbf{The EMA teacher prevents long-horizon collapse.}
Training \methodname{} with the teacher scored by the current policy (no EMA) vs.\ the EMA teacher ($\alpha{=}0.99$), otherwise identical. Left: policy entropy. Middle: reference-policy KL. Right: training reward averaged across image and text modalities. The current-policy teacher destabilizes after ${\sim}90$ steps; the EMA teacher stays stable and reaches higher reward.}
\label{fig:ema_teacher_stability}

\end{figure}

\textcolor{lightblue}{\textbf{Ablation 2: a slow-moving EMA teacher is necessary for long-horizon stability.}}
Figure~\ref{fig:ema_teacher_stability} compares \methodname{} using an EMA teacher ($\alpha=0.99$) with a variant using the current policy $\theta$ as the teacher, under otherwise identical settings ($\lambda_{\text{KL}}=0.01$). The runs behave similarly for the first $50$ steps, after which the current-policy teacher exhibits self-reinforcing drift: policy entropy and reference-policy KL increase while training reward collapses. In contrast, the EMA teacher remains stable. We interpret this difference as a moving-target effect analogous to the instability addressed by target networks in value-based RL~\citep{mnih2015human}: decoupling the target's timescale from the student's stabilizes optimization without sacrificing asymptotic performance.

%% file: sections/conclusion_arxiv.tex
\section{Discussion and Perspectives on Future Work}
\label{conclusion}

In this work, we identify reasoning asymmetry across modalities as a source of supervision signal for multimodal reasoning. We introduce \methodfullname{} (\methodname{}), which selects the strongest-performing view of each problem as an internal teacher and regularizing the student-view policy toward it during RL. Our results show that paired multimodal data is most effective when the training objective explicitly exploits the correspondence between views. Simply mixing text-dominant and image-dominant views into the RL distribution does not reliably transfer reasoning, because each view is still optimized as an independent prompt. In contrast, \methodname{} makes the transfer direction explicit: the empirically stronger view provides guidance for the view being trained. This yields stronger and more consistent reasoning, with the largest gains on examples exhibiting pronounced cross-view performance differences.

More broadly, our findings suggest a different way to use multimodal data. Rather than treating different input formats as independent examples, we can treat them as alternative views of the same latent problem and use their disagreements to expose missing reasoning capabilities. This perspective opens several directions for future work. Extending beyond geometry and charts to scientific figures, tables, robotics observations, or broader visual reasoning will require reliable ways to align views by problem identity and answer target while accounting for differences in their information content. Teacher selection could also be improved with uncertainty-aware weighting or alignment at intermediate reasoning steps, rather than relying only on rollout rewards. Overall, our findings point to a broader opportunity: cross-modal inconsistency can be transformed from a robustness concern into a useful training signal for learning multimodal reasoners that are stronger and less sensitive to how information is represented.

%% file: sections/acknowledgements.tex
\vspace{-0.2cm}
\section{Acknowledgements}
\vspace{-0.2cm}

YQ gratefully acknowledges the support of the Amazon AI PhD Fellowship.

%% file: sections/app_additional_results.tex
\section{Controlled Single-Problem Analysis and Online Teacher Dynamics}
\label{app:single_problem_analysis}
\label{app:teacher_dynamics}

Given the cross-view asymmetries identified in Section~\ref{sec:modality_diff}, two related questions arise. First, if standard RL observes text-dominant and image-dominant views of the same problem, will successful reasoning automatically transfer from the stronger view to the weaker one? Second, can \methodname{} provide useful guidance when the strongest view has only a small absolute success rate?

Teacher selection in \methodname{} is based on relative empirical success, so the selected view is not necessarily an accurate oracle. For example, the strongest view may solve only $2$ of $16$ rollouts while the other two views solve none. \methodname{} does not require the selected teacher to be highly accurate; it requires only a relative capability advantage over the student view. Unlike supervised fine-tuning on teacher-generated solutions, \methodname{} does not train the student to imitate teacher trajectories. The student samples its own trajectories, and the selected teacher distribution provides token-level reverse-KL guidance on those student-generated states. At the same time, the outcome-reward RL objective continues to anchor learning to task correctness regardless of teacher strength. The student is therefore neither trained solely by the teacher nor bounded by the teacher's initial success rate.

\begin{figure}[h]
\centering
\includegraphics[width=0.55\linewidth]{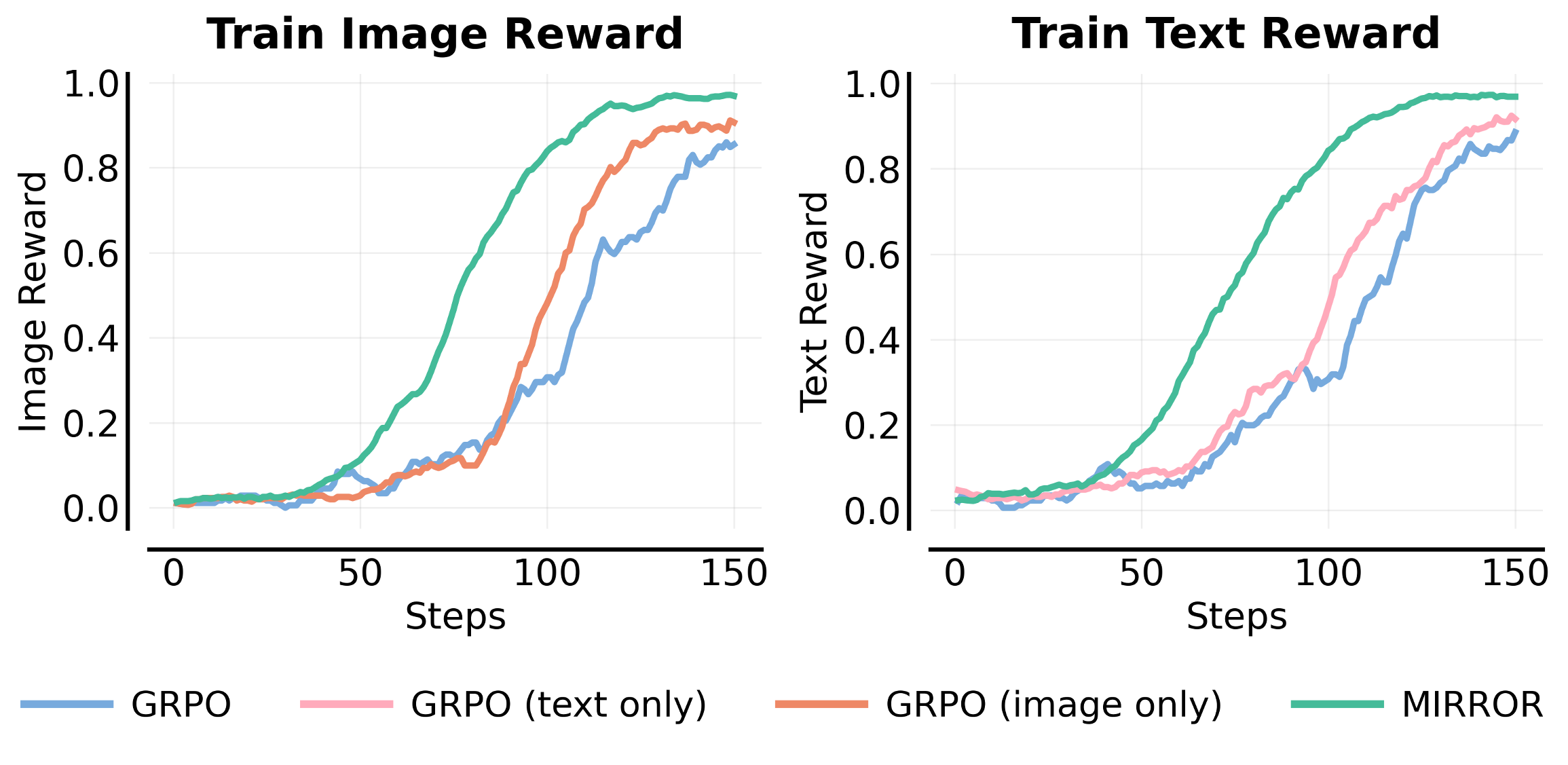}
\caption{\footnotesize
\textbf{Controlled single-problem analysis.} Across both views, \methodname{} achieves the highest training reward, followed by GRPO trained on the corresponding single modality, while naive GRPO over mixed-modality views performs worst. Simply exposing RL to multiple views of the same underlying problem is therefore insufficient for reliable cross-modal transfer; without an explicit transfer direction, the sparse reward signal can be diluted across heterogeneous problem--view pairs.}
\label{fig:single_problem_analysis}
\end{figure}

We study these questions through a controlled single-problem experiment comparing mixed-modality GRPO, text-only GRPO, image-only GRPO, and \methodname{}. In the mixed-modality condition, the same underlying problem appears under both text-dominant and image-dominant views, but each view is treated as an independent RL prompt and transfer must emerge implicitly through shared policy updates. The single-view conditions remove this cross-view heterogeneity and optimize only the corresponding student view. Figure~\ref{fig:single_problem_analysis} shows the training-reward trajectories, while Table~\ref{tab:weak_teacher_bootstrap} reports endpoint success counts from the same experiment before training and after 100 optimization steps.

Initially, the text-dominant view has only a small advantage over the image-dominant view ($1/16$ versus $0/16$). Nevertheless, Figure~\ref{fig:single_problem_analysis} and Table~\ref{tab:weak_teacher_bootstrap} show that \methodname{} reaches $13/16$ successful rollouts under both views after 100 steps, compared with $5/16$ under mixed-modality GRPO and $8/16$ under the corresponding single-view GRPO runs. Thus, the student is not bounded by the teacher view's initial performance, both views can improve as teacher selection and policy optimization are updated online, and directed cross-view guidance accelerates this co-improvement relative to mixed-modality and single-view RL. In contrast, naive mixed-modality GRPO obtains the lowest reward on both views, showing that exposure to paired views alone does not reliably transfer successful reasoning without an explicit direction of guidance.

\begin{table}[h]
\centering
\caption{\footnotesize
\textbf{Endpoint success counts for the controlled experiment in Figure~\ref{fig:single_problem_analysis}.}
Each entry reports successful rollouts out of 16 on the same problem before training and after 100 optimization steps. Dashes denote views excluded from the corresponding single-view GRPO run.}
\label{tab:weak_teacher_bootstrap}
\small
\setlength{\tabcolsep}{10pt}
\renewcommand{\arraystretch}{1.08}
\begin{tabular}{@{}l c c c c@{}}
\toprule
& \multicolumn{2}{c}{\textbf{Image-dominant}} & \multicolumn{2}{c}{\textbf{Text-dominant}} \\
\cmidrule(lr){2-3}\cmidrule(lr){4-5}
\textbf{Setting} & \textbf{Step 0} & \textbf{Step 100} & \textbf{Step 0} & \textbf{Step 100} \\
\midrule
Mixed-modality GRPO & $0/16$ & $5/16$ & $1/16$ & $5/16$ \\
GRPO (text only)    & --     & --     & $1/16$ & $8/16$ \\
GRPO (image only)   & $0/16$ & $8/16$ & --     & --     \\
\methodname{}       & $0/16$ & $\mathbf{13/16}$ & $1/16$ & $\mathbf{13/16}$ \\
\bottomrule
\end{tabular}
\end{table}

The relative advantages used for teacher selection are typically larger than the one-rollout gap in the controlled example. Before training, the median difference between the strongest and weakest views across problems is $0.2188$, corresponding to $7/32$ rollouts. The mean success rates of the strongest, second-strongest, and weakest views are $0.3602$, $0.2348$, and $0.0751$, respectively. Thus, although the selected teacher is not assumed to be an oracle, the initial views commonly exhibit a nontrivial relative capability gap that can provide directional guidance. These initial gaps form the starting point for online teacher selection, but the assignment need not remain fixed as the model learns.

Because \methodname{} performs teacher selection online, the selected view can change as text-dominant, image-dominant, and combined-view capabilities evolve. Table~\ref{tab:teacher_selection_dynamics} compares the teacher-selection proportions averaged over the first and last 50 training steps. \emph{No teacher} denotes cases in which no alternative view performs better than the current student view and therefore no view offers a relative capability advantage for cross-view guidance.

\begin{table}[h]
\centering
\caption{\footnotesize
\textbf{Online teacher-selection dynamics.}
We report the mean teacher-selection percentage and its variation over the first and last 50 training steps.}
\label{tab:teacher_selection_dynamics}
\small
\setlength{\tabcolsep}{14pt}
\renewcommand{\arraystretch}{1.08}
\begin{tabular}{@{}l c c@{}}
\toprule
\textbf{Teacher view} & \textbf{Early training} & \textbf{Late training} \\
\midrule
Combined   & $29.67 \pm 3.31$ & $28.17 \pm 2.88$ \\
Text       & $18.05 \pm 2.19$ & $14.23 \pm 1.86$ \\
Image      & $ 5.30 \pm 1.26$ & $ 5.25 \pm 1.30$ \\
No teacher & $46.98 \pm 2.64$ & $52.35 \pm 2.64$ \\
\bottomrule
\end{tabular}
\end{table}

The distribution changes moderately rather than abruptly. The combined view remains the most frequently selected teacher, decreasing slightly from $29.67\%$ to $28.17\%$. Text-teacher selection decreases from $18.05\%$ to $14.23\%$, while image-teacher selection remains nearly unchanged. Meanwhile, the no-teacher proportion increases from $46.98\%$ to $52.35\%$. This increase indicates that, as training progresses, a larger fraction of student views no longer has an alternative view that performs better, so fewer examples offer a relative advantage for cross-view guidance. The changing proportions also show that modality dominance is not fixed by the base model, supporting online, problem-specific teacher selection rather than assigning one teacher modality before training.

\section{Generalization to Larger Base Models}
\label{app:qwen8b}

The main experiments use Qwen3-VL-4B-Instruct. To test whether the benefits of cross-view supervision are specific to a single base model or scale, we repeat the comparison with Qwen3-VL-8B-Instruct~\citep{bai2025qwen3vltechnicalreport} and GLM-4.1V-9B-Thinking~\citep{glm-vteam2025glm41vthinking}. For each model, we compare the base model with GRPO and \methodname{}. Table~\ref{tab:qwen8b_results} reports pass@1 and pass@16 on ODA-Val, image- and text-view pass@1 on ODA-Unfiltered, pass@1 and pass@16 on GeoInt, and the official five-split mean judge score on MathVerse.

\begin{table}[h]
\centering
\caption{\footnotesize
\textbf{\methodname{} generalizes to larger base models.}
Within each model family, \methodname{} improves over the base model on every reported metric. ODA-Unfiltered contains 309 problems and is evaluated with 32 rollouts per problem and view. MathVerse values are averages over the five official splits.}
\label{tab:qwen8b_results}
\small
\setlength{\tabcolsep}{5.5pt}
\renewcommand{\arraystretch}{1.05}
\resizebox{\linewidth}{!}{%
\begin{tabular}{@{}l c c c c c c c c c@{}}
\toprule
\textbf{Approach}
& \multicolumn{2}{c}{\textbf{ODA-Val Image}}
& \multicolumn{2}{c}{\textbf{ODA-Val Text}}
& \multicolumn{2}{c}{\textbf{ODA-Unfiltered}}
& \multicolumn{2}{c}{\textbf{GeoInt}}
& \textbf{MathVerse} \\
\cmidrule(lr){2-3}\cmidrule(lr){4-5}\cmidrule(lr){6-7}\cmidrule(lr){8-9}\cmidrule(lr){10-10}
& pass@1 & pass@16 & pass@1 & pass@16 & \shortstack{image\\pass@1} & \shortstack{text\\pass@1} & pass@1 & pass@16 & mean \\
\midrule
Qwen3-VL-8B-Instruct
& 14.40 & 45.79
& 34.48 & 82.41
& 37.0 & 37.7
& 58.81 & 74.07
& 55.30 \\
GRPO
& 17.29 & 41.73
& 36.10 & 75.81
& 30.1 & 33.5
& 65.41 & 79.67
& 55.81 \\
\methodname{}
& \textbf{26.49} & \textbf{60.90}
& \textbf{48.87} & \textbf{87.73}
& \textbf{41.7} & \textbf{43.5}
& \textbf{68.47} & \textbf{80.08}
& \textbf{57.54} \\
\midrule
GLM-4.1V-9B-Thinking
& 14.18 & 34.61
& 25.94 & 57.99
& 25.6 & 27.4
& 56.70 & 73.30
& 69.65 \\
GRPO
& 19.50 & 40.20
& 31.74 & 61.51
& 29.0 & 29.6
& 62.60 & 76.40
& 70.03 \\
\methodname{}
& \textbf{24.54} & \textbf{49.98}
& \textbf{33.75} & \textbf{63.19}
& \textbf{29.5} & \textbf{30.0}
& \textbf{64.10} & \textbf{78.50}
& \textbf{70.36} \\
\bottomrule
\end{tabular}%
}
\end{table}

For Qwen3-VL-8B-Instruct, standard GRPO improves GeoInt and ODA-Val pass@1, but reduces ODA-Val image pass@16 from $45.79\%$ to $41.73\%$, text pass@16 from $82.41\%$ to $75.81\%$, and both ODA-Unfiltered pass@1 scores. In contrast, \methodname{} improves all nine reported metrics over the base model. Its gains are largest on ODA-Val, where image pass@1 and pass@16 improve by $12.09$ and $15.11$ pp, and text pass@1 and pass@16 improve by $14.39$ and $5.32$ pp. It also improves ODA-Unfiltered image/text pass@1 by $4.7$/$5.8$ pp, GeoInt pass@1/pass@16 by $9.66$/$6.01$ pp, and the MathVerse mean judge score by $2.24$ points.

The GLM-4.1V-9B-Thinking results show the same pattern. Relative to its base model, \methodname{} improves ODA-Val image pass@1/pass@16 by $10.36$/$15.37$ pp, ODA-Val text pass@1/pass@16 by $7.81$/$5.20$ pp, ODA-Unfiltered image/text pass@1 by $3.9$/$2.6$ pp, GeoInt pass@1/pass@16 by $7.40$/$5.20$ pp, and the official five-split MathVerse mean by $0.71$ points. It also exceeds GRPO on all nine reported metrics. Together, these results show that the cross-view training signal is not limited to the 4B base model used in the main experiments or to a single larger-model family.

\section{Generalization to a New Domain: Chart Reasoning}
\label{app:chart_generalization}

We further test whether the cross-view training framework generalizes beyond geometry by applying it in the chart domain. ChartX~\citep{xia2024chartxchartvlm} is a multimodal chart-reasoning benchmark spanning 18 chart types, 7 tasks, and 22 disciplinary topics. Each chart is represented through four modalities: a chart image, its underlying CSV data, the corresponding Python code, and a text description. In this experiment, we use only the question-answering task. We construct the chart-domain training data in the same manner as ODA-Data: the image and CSV representations form paired views of the same question and underlying chart information, and we select examples that exhibit view-dependent solvability asymmetry under the base model. Starting from Qwen3-VL-4B, we post-train on these asymmetric paired question-answering examples with either standard GRPO or \methodname{}, and evaluate pass@1 accuracy for both input representations.

\begin{table}[h]
\centering
\caption{\footnotesize
\textbf{Cross-view training generalizes to chart reasoning.}
Results are pass@1 accuracy on the ChartX question-answering task after chart-domain post-training on image and CSV views. The final column reports the improvement of \methodname{} over standard GRPO in pp.}
\label{tab:chart_generalization}
\small
\setlength{\tabcolsep}{10pt}
\renewcommand{\arraystretch}{1.05}
\begin{tabular}{@{}l c c c c@{}}
\toprule
\textbf{Input view} & \textbf{Base} & \textbf{GRPO} & \textbf{\methodname{}} & \textbf{$\Delta$ vs. GRPO} \\
\midrule
Image & 58.49 & 61.71 & \textbf{63.74} & +2.03 \\
CSV   & 82.41 & 92.35 & \textbf{93.15} & +0.80 \\
\bottomrule
\end{tabular}
\end{table}

Both chart-domain post-training methods improve over the base model, and \methodname{} is strongest under both chart representations. Relative to the base model, \methodname{} improves pass@1 by $5.25$ pp for chart images and $10.74$ pp for CSV inputs. It also outperforms standard GRPO by $2.03$ pp on images and $0.80$ pp on CSV inputs. The larger advantage on the image view suggests that cross-view supervision is particularly helpful when the model must recover values and relations from a visual representation rather than reading them directly from structured text. Together with the geometry results, this experiment shows that the training framework can be instantiated in a distinct domain with different input representations.

\section{Consistency of Cross-View Asymmetry Across Model Scales}
\label{app:cross_model_asymmetry}

The effectiveness of \methodname{} depends on different views exposing different capabilities of the current model. We therefore examine whether the direction of this asymmetry is stable when model scale changes. On the \methodname{} training set, we sample 32 rollouts per view from both Qwen3-VL-4B-Instruct and Qwen3-VL-8B-Instruct. For problem $i$ and model $m$, let $\hat p^{(m)}_{i,\mathrm{image}}$ and $\hat p^{(m)}_{i,\mathrm{text}}$ denote the empirical image-view and text-view success rates. We define the signed modality gap as
\begin{align}
g_i^{(m)} = \hat p^{(m)}_{i,\mathrm{image}} - \hat p^{(m)}_{i,\mathrm{text}}.
\end{align}
A positive gap indicates that the image view is stronger, a negative gap indicates that the text view is stronger, and a zero gap is a tie. We call a problem strictly asymmetric when $g_i^{(m)} \neq 0$.

\begin{table}[h]
\centering
\caption{\footnotesize
\textbf{Cross-view asymmetry persists across model scales.}
Preference categories are computed from 32 rollouts per view on the \methodname{} training set. Percentages are rounded independently.}
\label{tab:cross_model_asymmetry}
\small
\setlength{\tabcolsep}{12pt}
\renewcommand{\arraystretch}{1.05}
\begin{tabular}{@{}l c c@{}}
\toprule
\textbf{Image--text preference} & \textbf{Qwen3-VL-4B} & \textbf{Qwen3-VL-8B} \\
\midrule
Image view stronger & 15.7\% & 18.1\% \\
Text view stronger & 70.4\% & 70.5\% \\
Tie & 13.8\% & 11.3\% \\
Strict cross-view asymmetry & 86.2\% & 88.7\% \\
\bottomrule
\end{tabular}
\end{table}

The aggregate preference distribution is similar at both scales. The text view is stronger for roughly $70\%$ of examples for each model, while strict cross-view asymmetry occurs for $86.2\%$ of examples at 4B and $88.7\%$ at 8B. The similarity is also visible at the problem level. Among examples for which both models have a strict preference, the models agree on the stronger view in $89.6\%$ of cases. Moreover, their signed modality gaps are strongly correlated, with Pearson correlation $r=0.779$. Thus, scaling the model from 4B to 8B does not eliminate view-dependent solvability or substantially change its overall direction.

At the same time, preference is not completely invariant to model scale. In the remaining $10.4\%$ of examples where both models have a strict preference, one model favors the image view while the other favors the text view. A teacher assignment computed once using a different model or checkpoint would therefore be incorrect for a non-negligible subset of examples. This residual disagreement supports the online, model-specific teacher selection in \methodname{}, which continually chooses the stronger view according to the capabilities of the model being trained rather than relying on a fixed cross-model preference label.

%% file: sections/app_error_analysis.tex
\section{Error Analysis of Cross-View Transfer}
\label{app:error_analysis}

To better understand what \methodname{} transfers across modalities, we conduct an error analysis on a random sample comprising $15\%$ of the data used by \methodname{}. For both Qwen3-VL-4B-Instruct and \methodname{}, we generate 32 rollouts per problem under each of the text-dominant and image-dominant views. A Gemini judge classifies the first incorrect response from each model into one of three primary error categories: perception, reasoning, or arithmetic/final-answer. When a model solves all 32 rollouts for a problem, we instead assign the outcome \emph{No error}.

\begin{table}[h]
\centering
\caption{\footnotesize
Primary error outcomes before and after \methodname{} training.}
\label{tab:error_analysis}
\small
\setlength{\tabcolsep}{7pt}
\renewcommand{\arraystretch}{1.08}
\resizebox{\linewidth}{!}{%
\begin{tabular}{@{}l c c c c@{}}
\toprule
& \multicolumn{2}{c}{\textbf{Text-dominant}} & \multicolumn{2}{c}{\textbf{Image-dominant}} \\
\cmidrule(lr){2-3}\cmidrule(lr){4-5}
\textbf{Primary outcome} & \textbf{Qwen3-VL-4B-Instruct} & \textbf{\methodname{}} & \textbf{Qwen3-VL-4B-Instruct} & \textbf{\methodname{}} \\
\midrule
Perception error              & 0.0\%  & 0.0\%  & 53.0\% & 51.7\% \\
Reasoning error               & 52.4\% & 44.8\% & 16.0\% & 10.7\% \\
Arithmetic/final-answer error & 47.5\% & 45.5\% & 31.0\% & 31.0\% \\
No error                      & 0.0\%  & 9.7\%  & 0.0\%  & 6.7\%  \\
\bottomrule
\end{tabular}%
}
\end{table}

The largest reduction is in reasoning errors: $7.6$ pp for text-dominant inputs and $5.3$ pp for image-dominant inputs. Moreover, among problems on which the base model produces an error, \methodname{} is error-free across all 32 sampled rollouts in $9.7\%$ of text-dominant cases and $6.7\%$ of image-dominant cases. In contrast, arithmetic/final-answer errors decrease only slightly for text-dominant inputs and remain unchanged for image-dominant inputs. These shifts suggest that \methodname{}'s improvements include mainly gains in reasoning rather than arising only from corrections to arithmetic or final-answer production.

%% file: sections/app_computation.tex
\section{FLOPs Estimation}
\label{app:compute}

We estimate the computational overhead of \methodname{} relative to standard mixed-modality GRPO.
Following the common approximation used in language-model scaling analyses, we count one forward-only pass over $D$ tokens as approximately $2ND$ FLOPs~\citep{sardana2025chinchillaoptimalaccountinginferencelanguage}, where $N$ is the number of model parameters, and one training pass with forward and backward propagation as approximately $6ND$ FLOPs~\citep{hoffmann2022trainingcomputeoptimallargelanguage}. This gives a coarse but useful estimate of the relative cost between rollout generation, teacher-logit evaluation, and gradient updates.

\paragraph{Setup.}
Standard mixed-modality GRPO trains on two restricted student views: text-dominant and image-dominant. For each problem, it samples $K$ rollouts from each view, giving $2K$ online student rollouts. \methodname{} uses the same two restricted views for student RL training, but performs teacher selection over three candidate views:
$$\mathcal{M}=\{\text{text},\text{img},\text{comb}\}$$

The text-dominant and image-dominant candidate rollouts are exactly the online rollouts already generated for GRPO, so they do not need to be recomputed. Therefore, the only additional rollout generation required by \methodname{} is the $K$ rollouts from the combined image+text view.

\paragraph{Mixed-modality GRPO cost.}
Let $D$ denote the average rollout length. Standard mixed-modality GRPO generates $2K$ rollouts per problem, so its forward-only rollout-generation cost is approximately
\begin{align}
C_{\mathrm{mixed\text{-}GRPO,gen}}
&\approx
2K \cdot 2ND
=
4KND.
\end{align}
It then performs one student-side policy update on these $2K$ rollouts. Using the standard training approximation, the forward-backward update cost is
\begin{align}
C_{\mathrm{mixed\text{-}GRPO,train}}
&\approx
2K \cdot 6ND
=
12KND.
\end{align}
Thus, ignoring implementation-specific constants, the total cost is
\begin{align}
C_{\mathrm{mixed\text{-}GRPO}}
&\approx
4KND + 12KND
=
16KND.
\end{align}

\paragraph{\methodname{} cost.}
\methodname{} reuses the same $2K$ text-dominant and image-dominant student rollouts as mixed-modality GRPO. It additionally samples $K$ combined-view rollouts for online teacher selection. Therefore, its rollout-generation cost is
\begin{align}
C_{\methodname,\mathrm{gen}}
&\approx
3K \cdot 2ND
=
6KND.
\end{align}
After selecting the teacher view, \methodname{} computes the reverse-KL regularizer on the $2K$ student-view rollouts. This requires a teacher-forced forward pass under the selected teacher view to compute teacher log-probabilities on the same generated tokens using the EMA teacher weights. Since this is forward-only and stop-gradient, its cost is approximately
\begin{align}
C_{\methodname,\mathrm{KL\text{-}eval}}
&\approx
2K \cdot 2ND
=
4KND.
\end{align}
The student-side GRPO update is unchanged from mixed-modality GRPO:
\begin{align}
C_{\methodname,\mathrm{train}}
&\approx
2K \cdot 6ND
=
12KND.
\end{align}
Therefore, the total cost of \methodname{} is approximately
\begin{align}
C_{\methodname}
&\approx
6KND + 4KND + 12KND
=
22KND.
\end{align}

\paragraph{Relative overhead.}
Under this coarse FLOPs model, the relative cost of \methodname{} compared with standard mixed-modality GRPO is
\begin{align}
\frac{C_{\methodname}}{C_{\mathrm{mixed\text{-}GRPO}}}
&\approx
\frac{22KND}{16KND}
=
1.375.
\end{align}
Thus, \methodname{} adds roughly $37.5\%$ FLOPs over standard mixed-modality GRPO under a single-update accounting. This overhead comes only from forward-only computation: $K$ additional combined-view rollouts for teacher selection and a teacher-logit evaluation pass for the reverse-KL regularizer. \methodname{} uses the same trainable model, the same $2K$ student-view RL rollouts, and the same number of backward updates as mixed-modality GRPO. To examine whether standard GRPO would obtain similar optimization gains under comparable compute, we inspect the running-average training rewards at \methodname{} step 150 and GRPO step 200 shown in figure ~\ref{fig:mixed_modality_transfer}, which correspond to approximately similar cumulative FLOPs. At these points in training, \methodname{} has higher image reward ($0.2625$ vs.\ $0.2368$) and text reward ($0.4581$ vs.\ $0.4387$). This suggests that the advantage is not solely due to the additional per-update computation.

\paragraph{EMA teacher overhead.}
The EMA teacher does not change the FLOPs accounting above: the teacher-logit evaluation pass ($C_{\methodname,\mathrm{KL\text{-}eval}}\approx 4KND$) is simply executed with the EMA weights $\bar\theta$ instead of the current weights $\theta$, and the Polyak update in Equation~\ref{eq:ema_update} is a single $O(N)$ elementwise operation per training step (under one second per step in our runs, versus roughly $20$ minutes per full training step). The main cost is memory: one additional forward-only copy of the policy parameters, kept in fp32 for numerically stable high-decay averaging and offloaded to CPU between teacher evaluations. Checkpoints additionally store the EMA parameters so that training resumes exactly.

%% file: sections/app_pseudocode.tex
\section{Pseudocode for \methodname{}}

Algorithm~\ref{alg:mirror} outlines the training loop of \methodname{}.

\label{sec:app_code}
\begin{algorithm}[ht]
\caption{\methodname{} update loop}
\label{alg:mirror}
\begin{algorithmic}[1]
\Require policy $\pi_\theta$, frozen reference policy $\pi_{\mathrm{ref}}$, EMA teacher decay $\alpha$, paired dataset $\mathcal{D}$, rollouts per prompt $K$, reverse-KL coefficient $\lambda_{\mathrm{KL}}$, learning rate $\eta$
\Statex
\State \textbf{Dataset.} Each problem has paired text-dominant $x^{\mathrm{text}}$ and image-dominant $x^{\mathrm{image}}$ views, sharing the same pair identifier. Each row also has a fixed combined image+text input $x^{\mathrm{comb}}$ precomputed at dataset construction time.
\State \textbf{Sampler.} Construct minibatches so that paired rows from the same problem appear in the same minibatch and on the same data-parallel rank.
\State \textbf{EMA teacher.} Initialize EMA parameters $\bar\theta \gets \theta$.
\Statex
\For{each epoch}
    \For{each minibatch $B$}
        \State Set $\pi_{\mathrm{old}} \gets \pi_\theta$.

        \Statex
        \State \textbf{Standard rollouts from student views.}
        \For{each row $x \in B$}
            \State Sample $N$ responses $\{y_i\}_{i=1}^{N} \sim \pi_{\mathrm{old}}(\cdot \mid x)$.
            \State Compute rewards $r_i \gets R(x,y_i)$ for $i=1,\ldots,K$.
        \EndFor

        \Statex
        \State \textbf{combined view rollouts.}
        \For{each row $x \in B$}
            \State Sample $N$ responses $\{y^{\mathrm{comb}}_j\}_{j=1}^{K} \sim \pi_{\mathrm{old}}(\cdot \mid x^{\mathrm{comb}})$.
            \State Compute rewards $r^{\mathrm{comb}}_j \gets R(x^{\mathrm{comb}},y^{\mathrm{comb}}_j)$ for $j=1,\ldots,K$.
        \EndFor

        \Statex
        \State \textbf{Teacher selection.}
        \For{each row $x \in B$}
            \State Let $x^{\mathrm{pair}}$ be the paired view of the same problem.
            \State Let $\mathcal{T}(x)$ contain three candidates: the current rollout $\pi_{\mathrm{old}}(\cdot \mid x)$, the paired-view rollout $\pi_{\mathrm{old}}(\cdot \mid x^{\mathrm{pair}})$ and the combined-view rollout $\pi_{\mathrm{old}}(\cdot \mid x_{\mathrm{comb}})$.
            \State Score each candidate using its rollout rewards.
            \State Select the highest-scoring valid candidate view $x^{\mathrm{sel}}$; define the teacher as the EMA model under that view: $\pi_T(\cdot \mid x) \gets \pi_{\bar\theta}(\cdot \mid x^{\mathrm{sel}})$.
            \State If the selected teacher is the current view, exclude $x$ from the reverse-KL loss.
        \EndFor

        \Statex
        \State \textbf{GRPO loss.}
        \State Compute GRPO advantages from the standard rollout rewards $\{r_i\}_{i=1}^{K}$.
        \State Compute $\mathcal{L}_{\mathrm{GRPO}}$ using $\pi_\theta$, $\pi_{\mathrm{old}}$, $\pi_{\mathrm{ref}}$, and the standard rollouts.

        \Statex
        \State \textbf{Reverse-KL loss.}
        \State Compute $\mathcal{L}_{\mathrm{rKL}}$ using the sampled K3 surrogate described after Equation~\ref{eq:rkl_obj}.

        \Statex
        \State \textbf{Update.}
        \State $\mathcal{L}_{\mathrm{total}} \gets \mathcal{L}_{\mathrm{GRPO}} + \lambda_{\mathrm{KL}}\mathcal{L}_{\mathrm{rKL}}$.
        \State Update $\theta$ by minimizing $\mathcal{L}_{\mathrm{total}}$ with learning rate $\eta$. 
        \State $\bar\theta \gets \alpha\,\bar\theta + (1-\alpha)\,\theta$ \Comment{EMA teacher update}
    \EndFor
\EndFor

\end{algorithmic}
\end{algorithm}

%% file: sections/app_hyperparameters.tex
\section{Training Details}
\label{app:training_details}

We train all models using GRPO with verl~\citep{sheng2025hybridflow}. Unless otherwise specified, experiments use \texttt{Qwen3-VL-4B-Instruct} as the base model, vLLM as the rollout engine, and $8$ nodes with $8$ H200 GPUs per node. 
Each training run uses a maximum prompt length of $4096$ tokens and a maximum response length of $16{,}384$ tokens. 
We use a sampling temperature of $0.8$ and generate $n=16$ rollouts per prompt. 
The training batch size is $256$, with PPO mini-batch size $128$ and micro-batch size $1$ per GPU. 
We use GRPO with clipping thresholds $\epsilon_{\mathrm{low}}=0.2$ and $\epsilon_{\mathrm{high}}=0.26$. 
We train with learning rate $1\times 10^{-6}$, entropy coefficient $0.001$, and a reference-policy KL loss coefficient of $0.001$ using the low-variance KL estimator. 
For \methodname{}, we additionally enable the cross-modal reverse-KL regularizer with coefficient $\lambda_{\mathrm{KL}}=0.01$ by default, and compute the teacher target log-probabilities from an EMA copy of the policy with decay $\alpha=0.99$, updated after every actor step. The EMA copy is kept in fp32 and is forward-only (no optimizer state); it is saved with each checkpoint so training resumes exactly.

Each RL training job uses $8$ nodes with $8$ H200 GPUs per node, runs at approximately $20$ minutes per training step, and is trained for at least $200$ steps, corresponding to roughly $67$ wall-clock hours or about $4{,}267$ H200 GPU-hours per model.

\begin{table}[h]
\vspace{-0.2cm}
\centering
\caption{\footnotesize \textbf{Main training hyperparameters.}
Unless otherwise specified, all \methodname{} and GRPO baselines use the same training configuration. 
The only method-specific differences are whether the cross-modal reverse-KL regularizer is enabled and its EMA teacher (decay $0.99$).}
\vspace{0.2cm}
\small
\setlength{\tabcolsep}{7pt}
\renewcommand{\arraystretch}{1.05}
\resizebox{0.7\linewidth}{!}{%
\begin{tabular}{@{}l l@{}}
\toprule
\textbf{Category} & \textbf{Setting} \\
\midrule
Base model & \texttt{Qwen/Qwen3-VL-4B-Instruct} \\
Training framework & verl~\citep{sheng2025hybridflow} \\
RL algorithm & GRPO \\
Rollout engine & vLLM \\
Maximum prompt length & $4096$ tokens \\
Maximum response length & $16{,}384$ tokens \\
Rollouts per prompt & $16$ \\
Sampling temperature & $0.8$ \\
Train batch size & $256$ \\
PPO mini-batch size & $128$ \\
PPO micro-batch size per GPU & $1$ \\
Learning rate & $1\times 10^{-6}$ \\
Clip ratio low & $0.2$ \\
Clip ratio high & $0.26$ \\
Entropy coefficient & $0.001$ \\
Reference KL loss & Enabled \\
Reference KL coefficient & $0.001$ \\
Reference KL type & Low-variance KL \\
Total epochs & $200$ \\
Cross-modal reverse-KL coefficient $\lambda_{\mathrm{KL}}$ & $0.01$ (\methodname{} only) \\
EMA teacher decay $\alpha$ & $0.99$ (\methodname{} only) \\
EMA teacher precision & fp32, forward-only \\
\bottomrule
\end{tabular}%
}
\label{tab:training_hyperparams}
\vspace{-0.3cm}
\end{table}

%% file: sections/app_prompt.tex
\newpage

\vspace{-0.2cm}
\section{ODA-Data Construction}
\label{sec:prompting}
\vspace{-0.2cm}

To study modality-dependent reasoning while holding the underlying problem fixed, we construct \textbf{ODA-Data}, a paired multimodal geometry dataset in which each problem is represented in two aligned views: a text-dominant view and an image-dominant view. Starting from 97K geometry-style problems from OpenDataArena/ODA-Math-460k~\citep{cai2025opendataarena}, we first apply a difficulty filter that removes easy questions with annotated pass@5=1, yielding 16K candidate questions. For each candidate, we prompt an LLM to generate a diagram in TikZ, a LaTeX-based language for programmatic scientific graphics. We choose TikZ over direct image generation because preliminary human inspection indicated that the model generated usable TikZ code more reliably and that the compiled diagrams had higher visual quality than directly generated images. We retain only diagrams whose TikZ code compiles successfully and use an LLM judge to verify that the generated diagram is semantically aligned with the original problem statement. We then apply a separate LLM judge to identify and filter out diagrams that leak the answer.

To construct the image-dominant view, we further prompt the judge to remove from the text any information already presented in the diagram, so that the remaining text and image provide complementary information. All LLM-based generation and judging steps use Gemini-3-Pro-Preview. We then perform modality-specific rollout evaluations using the base model, generating 32 rollouts for each of the text-dominant and image-dominant views, for a total of 64 rollouts per question. We define a view as failing when at most one of its 32 rollouts is correct. A question is retained only when exactly one of the two views meets this failure criterion; equivalently, one view has at most one successful rollout while the other has at least two. We exclude questions for which both views meet the failure criterion because they do not provide the cross-view advantage targeted by this construction. This filtering process yields approximately 2K curated paired multimodal examples.

We split these examples into \textbf{ODA-Train} and \textbf{ODA-Val} using an 85:15 split, with 15\% of the dataset held out for validation. \textbf{ODA-Train} is used for cross-modal post-training, while \textbf{ODA-Val} serves as a held-out benchmark for evaluating view-dependent failure, cross-modal consistency, and cross-view transfer. Each example contains the original text problem, generated diagram, image-dominant prompt, text-dominant prompt, answer annotations, and modality-specific rollout/evaluation metadata. This structure makes ODA-Data useful not only for training \methodname{}, but also for studying whether VLMs solve the same mathematical problem consistently across textual and visual representations.

For Tikz code generation, the prompt is:

\begin{promptbox}{Prompt for Tikz Code Generation}
\begingroup
$<$question$>$

Your task is to generate \textbf{only the TikZ (LaTeX) code} that draws the geometric diagram described in the problem.

Important rules:
\begin{itemize}
    \item Do NOT solve the problem.
    \item You may use colors (e.g., for clarity or emphasis), but:
    \begin{itemize}
        \item Do NOT use solid fills that completely cover underlying lines or shapes.
        \item If you use `fill', it must include an `opacity' value (e.g., `fill opacity=0.2') or be very light.
        \item Prefer `draw', `dashed', `thick', or colorized outlines over filled regions.
    \end{itemize}
    \item Avoid unnecessary decorations or stylistic choices not implied by the problem.
    \item Ensure the diagram is clear, uncluttered, and geometrically consistent.
\end{itemize}
Return only the TikZ code in a \emph{latex} block.
\endgroup
\end{promptbox}

For verifying the tikz compiled image against the question, the prompt is:

\begin{promptbox}{Prompt for Image Verification}
\begingroup
You are a strict geometry diagram judge.

You are given:
\begin{enumerate}
    \item A geometry problem described in text
    \item A diagram image intended to represent that problem
\end{enumerate}

Your task is to judge whether the diagram is an accurate and faithful representation of the problem text.

Check carefully that:
\begin{enumerate}
    \item All geometric entities mentioned in the text (points, lines, angles, shapes, objects) appear in the diagram.
    \item Spatial relationships match the text (relative positions, intersections, parallelism, perpendicularity, containment, motion if described).
    \item Angle markings are correct:
    \begin{itemize}
        \item Angles are marked at the correct vertex.
        \item The marked angle corresponds to the intended angle, not its complementary or supplementary angle.
        \item The rays defining the angle match the textual description.
    \end{itemize}
    \item Any stated angle measures or symbols (e.g., $90\circ$, $100\circ$, $\alpha$, $\beta$) are applied to the correct geometric region.
    \item The diagram does not introduce misleading constructions, extra constraints, or alternative interpretations that contradict the text.
    \item Do NOT judge based on visual scale accuracy:
    \begin{itemize}
        \item Differences in segment lengths, proportions, or drawing scale (e.g., a segment of length 15 not appearing longer than one of length 13) should NOT be treated as errors unless relative length ordering is explicitly required by the text.
    \end{itemize}
\end{enumerate}

If any required element is missing, incorrectly marked, ambiguously represented, or misleading, judge the diagram as inaccurate.

Respond with exactly the following format:

Judgement: Accurate or Inaccurate

Reason: A brief explanation (1–2 sentences) identifying the key correctness or primary issue.

Now evaluate the following.

[PROBLEM TEXT]

$<$QUESTION TEXT$>$

[DIAGRAM IMAGE]

$<$IMAGE$>$
\endgroup
\end{promptbox}

For rewriting image questions to remove information present in the diagram from the question, the prompt is:

\begin{promptbox}{Prompt for Text Rewrite}
\begingroup
You are given a math question that is accompanied by a diagram/image. Your task is to rewrite the question text so that information \textbf{already visible in the image} is removed from the text.

\#\# Rules:
\begin{enumerate}
    \item \textbf{Remove} textual descriptions of geometric configurations, diagrams, shapes, angles, side lengths, point positions, or any other information that a reader could extract from the image.
    \item \textbf{Only remove properties that are explicitly marked in the image.} For example, if the text says `isosceles triangle' and the diagram shows a triangle that look isosceles but does NOT have explicit markings (e.g., tick marks on equal sides, or a label saying `isosceles'), you must KEEP `isosceles triangle' in the text. The same applies to right angles (keep unless a square corner mark is shown), parallel lines (keep unless arrow marks are shown), equal segments (keep unless tick marks are shown), etc.
    \item \textbf{Always keep} the final question being asked (e.g., `Find the length of...', `What is the value of...', `Determine...').
    \item \textbf{Always keep} any answer format instructions (e.g., `put your final answer in \boxed{$\quad$}', `Final Answer: \boxed{$\quad$}').
    \item \textbf{Keep} any mathematical constraints or conditions that are purely algebraic and NOT depicted in the image.
    \item If the question is purely algebraic with no visual information to remove, return it UNCHANGED.
    \item Do NOT add any new information or rephrase the question -- only remove redundant parts.
    \item Return ONLY the rewritten question text, nothing else.
    \item Do not add placeholder tokens such as <image>.
\end{enumerate}

\#\# Original question:
$<$question$>$

\#\# Rewritten question (visual-redundant info removed):
\endgroup
\end{promptbox}

\section{Effect of Modality-Asymmetric Training Pairs}
\label{app:asymmetric_pairs}

As described in Section~\ref{sec:data_introduction}, ODA-Data contains problems that exhibit modality asymmetry. To test whether this asymmetry is important, we compare \methodname{} with a control model trained on the same number of properly paired examples that do not exhibit clear modality asymmetry, using identical hyperparameters. Training on asymmetric examples improves text-dominant validation pass@1 by $4.43$ pp and image-dominant validation pass@1 by $22.15$ pp relative to the paired-control set. These results show that paired representations alone are insufficient. The useful supervision arises when an empirically stronger view can provide problem-specific guidance to a view with very low rollout success, supporting our hypothesis that modality asymmetry is the primary source of self-supervision in \methodname{}.

%% file: sections/app_examples.tex
\section{Examples}
\label{app:examples}
\subsection{Example 1}

\begin{evalbox}{Summary}
\begingroup
The correct answer is $3\sqrt{2} + 2\sqrt{5}$. The image-dominant response realizes that P should sit on the intersection of $AC$ and $BD$ to minimize $|PA| + |PB| + |PC| + |PD|$ because this is intuitive from the diagram. The text-dominant response never realizes this and simply guess and exhausts different locations of P and only reaches a rough estimate of $|PA| + |PB| + |PC| + |PD| \approx 8.74$ 
\endgroup
\end{evalbox}
\begin{promptbox}{Image-dominant Prompt}
\begingroup
In the Cartesian coordinate plane, there are four fixed points $A, B, C, D$ and a moving point $P$. What is the minimum value of $|PA| + |PB| + |PC| + |PD|$?

\centering
\includegraphics[width=0.3\linewidth]{figures/no_p.png}

\endgroup
\end{promptbox}

\begin{solutionbox}{Image-dominant Response}
\begingroup
[...]

Step 25: Consider that the optimal point may be the intersection of the perpendicular bisectors or something else

[...]

Final Answer: $3\sqrt{2} + 2\sqrt{5}$ \cmark

\endgroup
\end{solutionbox}
\begin{promptbox}{Text-dominant Prompt}
\begingroup
In the Cartesian coordinate plane, there are four fixed points $A(-3,0)$, $B(1,-1)$, $C(0,3)$, $D(-1,3)$ and a moving point $P$. What is the minimum value of $|PA| + |PB| + |PC| + |PD|$?
\endgroup
\end{promptbox}

\begin{solutionbox}{Text-dominant Response }
\begingroup
[...]

Let me try P = (0, 3) — 9.36

Or try P = (0, 2) — 9.18

Or try P = (-1, 2) — 8.848

Or try P = (0, 2.5) — 8.756

Or try P = (-0.5, 2.5) — 8.756

Or try P = (-0.5, 2.4) — 8.744

Or try P = (-0.5, 2.34) — 8.741

So perhaps the minimum is 8.74.

Maybe the answer is 8.74 \xmark

\endgroup
\end{solutionbox}
\subsection{Example 2}

\begin{evalbox}{Summary}
\begingroup
The correct answer is 873. The text-dominant response successfully reasons through the set up of this question while the image-dominant response misreads the radii of each semicircle as half of the side length of the triangle. The text-dominant response realizes that for a semicircle with diameter on one side of the triangle and be tangent to the other two sides, its diameter must be less than the side of the triangle it sits on. 
\endgroup
\end{evalbox}

\begin{promptbox}{Image-dominant Prompt}
\begingroup
In a triangle, three semicircles can be drawn such that the diameter of each lies on a side of the triangle and each semicircle is tangent to the other two sides. The product of the areas of these three semicircles is numerically equal to the cube of the surface area of a certain sphere. The volume of this sphere can be expressed as $\dfrac{a\pi\sqrt{b}}{c}$, where $a$ and $c$ are relatively prime integers and $b$ is the product of distinct primes. Find $a+b+c$.

\centering
\includegraphics[width=0.3\linewidth]{figures/text_better.png}

\endgroup
\end{promptbox}

\begin{solutionbox}{Image-dominant Response }
\begingroup

Let’s consider the semicircle drawn on side $AB$ as diameter. Let $M$ be the midpoint of $AB$. The semicircle with diameter $AB$ has center $M$, radius $r = \frac{AB}{2} = 7.5$. For this semicircle to be tangent to sides $AC$ and $BC$, the distance from point $M$ to lines $AC$ and $BC$ must equal the radius $r = 7.5$.

[...]

So a + b + c = 455 + 2730 + 64 = 3249

3249 \xmark
\endgroup
\end{solutionbox}

\begin{promptbox}{Text-dominant Prompt}
\begingroup
In a triangle with side lengths $13$, $14$, and $15$, three semicircles can be drawn such that the diameter of each lies on a side of the triangle and each semicircle is tangent to the other two sides. The product of the areas of these three semicircles is numerically equal to the cube of the surface area of a certain sphere. The volume of this sphere can be expressed as $\dfrac{a\pi\sqrt{b}}{c}$, where $a$ and $c$ are relatively prime integers and $b$ is the product of distinct primes. Find $a+b+c$.
\endgroup
\end{promptbox}

\begin{solutionbox}{Text-dominant Response }
\begingroup
[...]

But wait — in a triangle, if you draw a semicircle with diameter on side AB, and you want it to be tangent to sides AC and BC, then the semicircle must lie **inside** the triangle, and its center lies on AB, and it touches AC and BC.

[...]

So a = 784, b = 2, c = 87, a + b + c = 784 + 2 + 87 = 873

So 873 \cmark

\endgroup
\end{solutionbox}

\subsection{Example 3}
\begin{evalbox}{Summary}
\begingroup
The correct answer is A. The privileged response correctly reaches the answer because it has all the necessary condition. The image-dominant response, though having all necessary condition, it did not read the $AP=AD$ condition from the diagram and reached the wrong asnwer B
\endgroup
\end{evalbox}
\begin{promptbox}{Image-dominant Prompt}
\begingroup
In the circle $O$, the length of $AB$ is twice the radius $R$. Then

(A) $AP^2=PB \cdot AB$.

(B) $AP \cdot DO=PB \cdot AD$.

(C) $AB^2=AD \cdot DE$.

(D) $AB \cdot AD=OB \cdot AO$.

(E) None of the above.

\centering
\includegraphics[width=0.3\linewidth]{figures/privileged_better.png}

\endgroup
\end{promptbox}

\begin{solutionbox}{Image-dominant Response }
\begingroup

We are given a circle with center $ O $, and a point $ A $ outside the circle such that $ AB $ is a tangent to the circle at point $ B $, and $ AB = 2R $, where $ R $ is the radius.

Also, $ AB \perp OB $, since a tangent is perpendicular to the radius at the point of contact. So triangle $ ABO $ is a right triangle with right angle at $ B $.

Moreover, $ D $ is a point on the circle and also on $ AO $, so $ D $ lies on the line segment from $ A $ to $ O $, and since $ O $ is the center, $ OD = R $ (radius), and since $ AO $ is a line from $ A $ to the center, and $ AB = 2R $, then $ AO $ is the hypotenuse of right triangle $ ABO $, with legs $ AB = 2R $ and $ OB = R $.

Let’s compute $ AO $:

In right triangle $ ABO $:

$$
AO = \sqrt{AB^2 + OB^2} = \sqrt{(2R)^2 + R^2} = \sqrt{4R^2 + R^2} = \sqrt{5R^2} = R\sqrt{5}
$$

Now, point $ D $ lies on $ AO $, and $ OD = R $, so since $ AO = R\sqrt{5} $, then $ AD = AO - OD = R\sqrt{5} - R = R(\sqrt{5} - 1) $.

We are to check which of the given options is true.

---

Let’s analyze each option one by one.

---

[...]

---

Therefore, the correct answer is \textbf{(B)}.

---

\textbf{Final Answer: \text{(B)}} \xmark

\endgroup
\end{solutionbox}

\begin{promptbox}{Combined Image+Text Prompt}
\begingroup
In the circle $O$, $BC$ is the diameter of the circle, and $AB \perp BC$. Also, $ADOE$ is a straight line, $AP=AD$, and the length of $AB$ is twice the radius $R$. Then

(A) $AP^2=PB \cdot AB$.

(B) $AP \cdot DO=PB \cdot AD$.

(C) $AB^2=AD \cdot DE$.

(D) $AB \cdot AD=OB \cdot AO$.

(E) None of the above.

\centering
\includegraphics[width=0.3\linewidth]{figures/privileged_better.png}

\endgroup
\end{promptbox}

\begin{solutionbox}{Combined Image+Text Response }
\begingroup

\textbf{Step 1: Understand the diagram and given data}
\begin{itemize}
    \item $BC$ is the diameter $\rightarrow$ so $BC = 2R$, and $O$ is the midpoint of $BC$.
    \item $AB \perp BC$ $\rightarrow$ so triangle $ABC$ is a right triangle at $B$.
    \item $ADOE$ is a straight line → points $A$, $D$, $O$, $E$ are colinear.
    \item $AP = AD$ → point $P$ lies on $AB$, and $AP = AD$.
    \item $AB = 2R$.
\end{itemize}

[...]

---

Therefore, only \textbf{(A)} is correct.

---

\textbf{Final Answer: $\text{(A)}$} \cmark
\endgroup
\end{solutionbox}
\subsection{Example 4}
\begin{evalbox}{Summary}
\begingroup
The image-dominant response wrongly reads the $5\sqrt{2}$ segment as AF instead of HF. Without the necessary and correct information, the image-dominant response never reaches a solution. 
\endgroup
\end{evalbox}
\begin{promptbox}{Image-dominant Prompt}
\begingroup
Given that $E$ is a point on the extension of side $BC$ of square $ABCD$, and a square $AEFG$ is constructed with $AE$ as one side. Then the length of $BE$ is?

\centering
\includegraphics[width=0.3\linewidth]{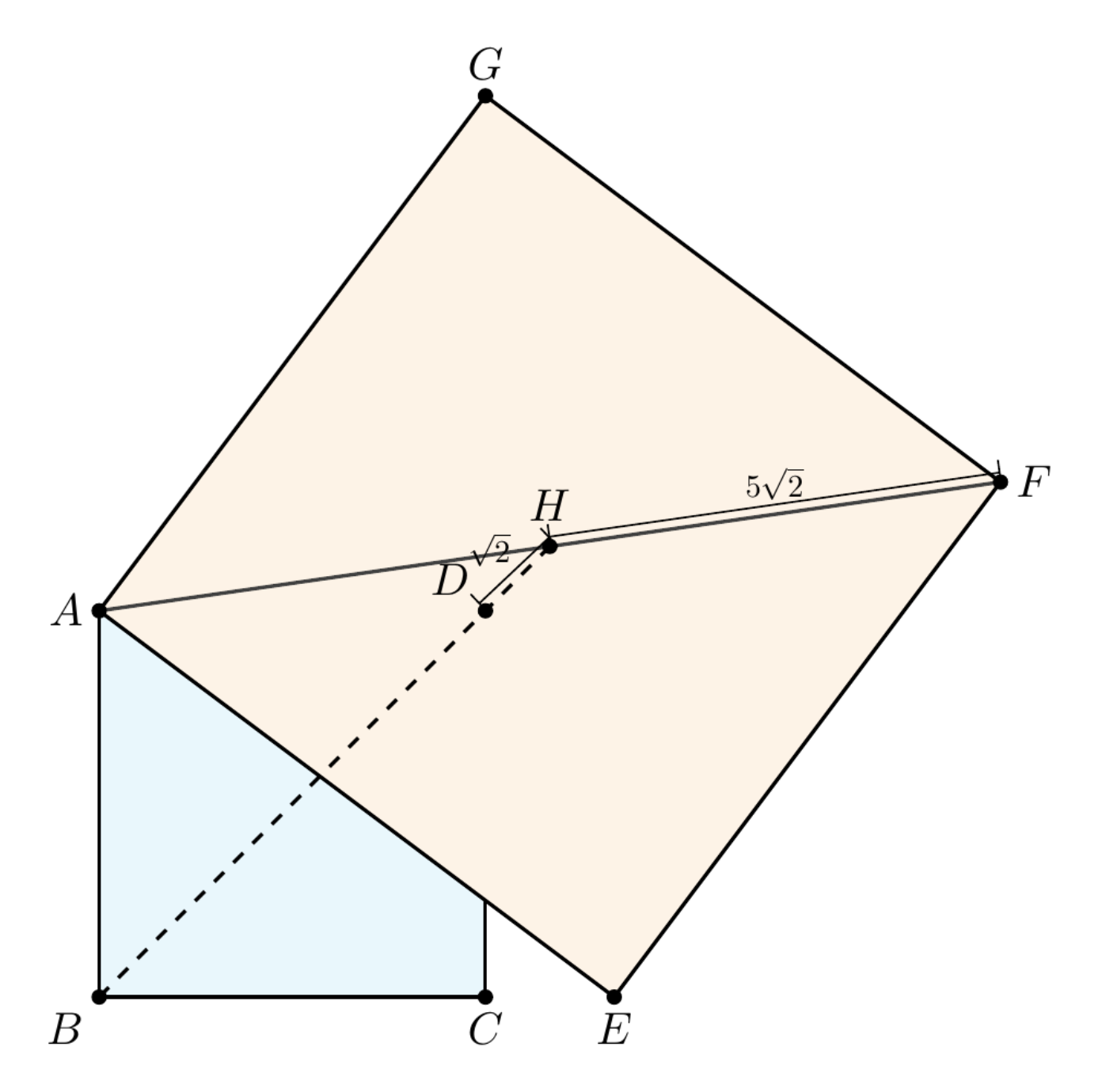}

\endgroup
\end{promptbox}

\begin{solutionbox}{Image-dominant Response}
\begingroup
We are given:

- Square $ABCD$, so all sides are equal and all angles are 90°.
- Point $E$ lies on the **extension** of side $BC$, meaning beyond point $C$.
- A square $AEFG$ is constructed with side $AE$.
- We are given the length $AF = 5\sqrt{2}$, and also the length $DH = \sqrt{2}$ (where $H$ is the point on $AF$ such that $DH$ is drawn and labeled as $\sqrt{2}$).

We are to find the length of $BE$.

[...]

\xmark

\endgroup
\end{solutionbox}
\begin{promptbox}{Text-dominant Prompt}
\begingroup
Given that $E$ is a point on the extension of side $BC$ of square $ABCD$, and a square $AEFG$ is constructed with $AE$ as one side, such that points $A$ and $G$ are on the same side of $BE$. The extension of $BD$ intersects $AF$ at point $H$, with $HD=\sqrt{2}$ and $FH=5\sqrt{2}$. Then the length of $BE$ is?
\endgroup
\end{promptbox}

\begin{solutionbox}{Text-dominant Response }
\begingroup

We are given a square $ABCD$, and point $E$ is on the extension of side $BC$. A square $AEFG$ is constructed with $AE$ as one side, and points $A$ and $G$ are on the same side of line $BE$. The extension of diagonal $BD$ intersects line $AF$ at point $H$, with $HD = \sqrt{2}$ and $FH = 5\sqrt{2}$. We are to find the length of $BE$.
[...]

Final Answer: $8$ \cmark

\endgroup
\end{solutionbox}

\subsection{Example 5}
\begin{evalbox}{Summary}
\begingroup
The text-dominant response treats 'Square A and B share a common side' as side length equal and could never solve the problem.  
\endgroup
\end{evalbox}
\begin{promptbox}{Text-dominant Prompt}
\begingroup
Square A, with side length 20 meters, has an area equal to the areas of square B and square C combined. Square A and square B share a common side. Square B and square C share a common side as well. What is the sum, in meters, of the combined perimeters of square B and square C?

\endgroup
\end{promptbox}

\begin{solutionbox}{Text-dominant Response}
\begingroup

We are given:

- Square A has side length **20 meters**, so its area is:
  $$
  \text{Area}_A = 20 \times 20 = 400 \text{ m}^2
  $$

- The area of square A equals the **combined areas** of square B and square C:
  $$
  \text{Area}_B + \text{Area}_C = 400
  $$

- Square A and square B **share a common side**, meaning they are adjacent — likely side by side or stacked, but since both are squares and share a side, **their side lengths must be equal**. Why? Because if two squares share a side, that side must have the same length for both squares. So:

  $$
  \text{Side length of square B} = \text{Side length of square A} = 20 \text{ meters}
  $$

Wait — but that would make area of B = 400 m², and then area of C would be 0, which is impossible.

So maybe they don't share a side of equal length? But the problem says “share a common side”. That implies that the side that is common must be of the same length for both squares.

So if square A and square B share a side, then they must both have that side as their side length. But then square B would also have side 20, so area = 400, and C would have area 0 — impossible.

Therefore, perhaps the common side is **not** the side of the square — that doesn’t make sense.
[...]

\xmark

\endgroup
\end{solutionbox}
\begin{promptbox}{Image-dominant Prompt}
\begingroup
Square A has an area equal to the areas of square B and square C combined. What is the sum, in meters, of the combined perimeters of square B and square C?

\centering
\includegraphics[width=0.3\linewidth]{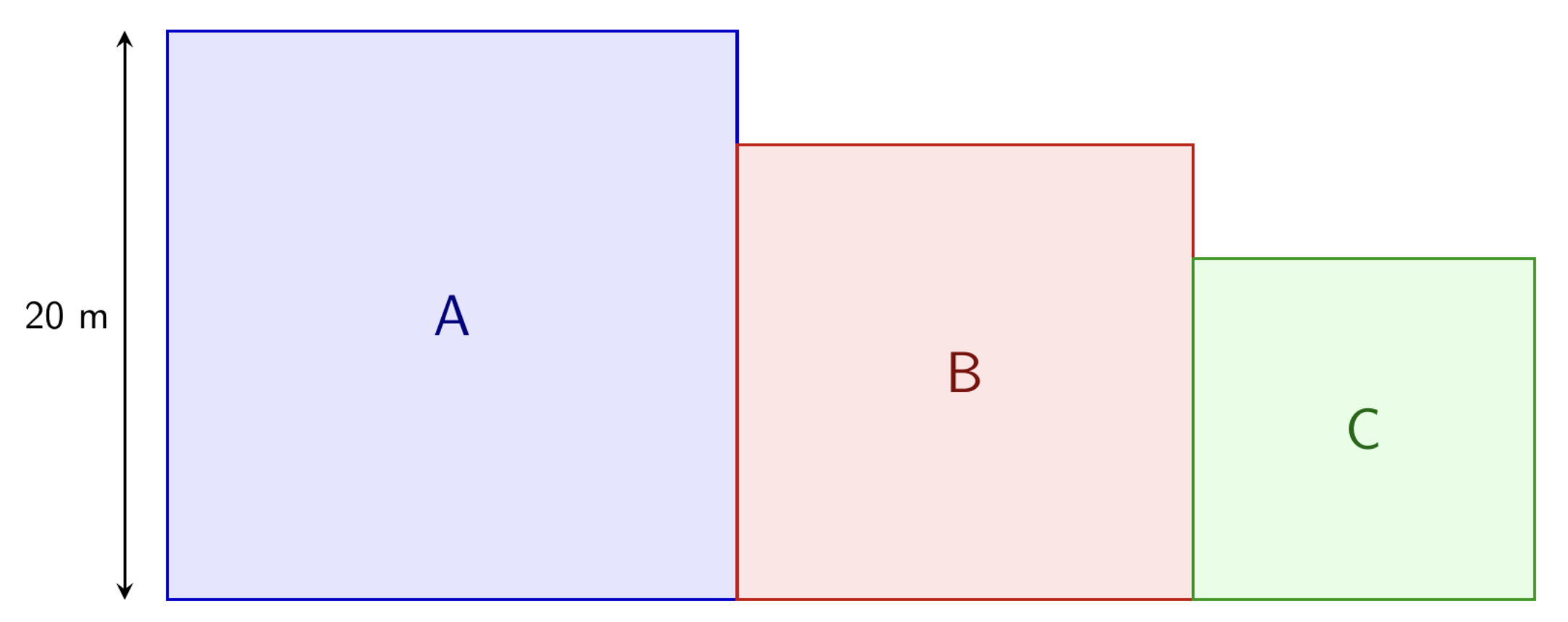}
\endgroup
\end{promptbox}

\begin{solutionbox}{Image-dominant Response }
\begingroup

We are given that Square A has an area equal to the combined areas of Square B and Square C. Also, the side length of Square A is given as **20 meters**.

---

Step 1: Find the area of Square A

[...]

Final Answer: $112$ meters \cmark

\endgroup
\end{solutionbox}